\documentclass[11pt]{article}
\usepackage{acl}

\usepackage{times}
\usepackage{latexsym}
\usepackage[T1]{fontenc}
\usepackage[utf8]{inputenc}
\usepackage{microtype}
\usepackage{inconsolata}
\usepackage{graphicx}
\usepackage{booktabs}
\usepackage{amsmath}
\usepackage{amssymb}
\usepackage{subcaption}
\usepackage[table]{xcolor}
\usepackage{multirow}
\usepackage{makecell}
\usepackage{enumitem}
\usepackage{colortbl}
\usepackage{placeins}

\definecolor{famQwen}{HTML}{2166ac}
\definecolor{famLLaVA}{HTML}{e08214}
\definecolor{famIntern}{HTML}{1b7837}
\definecolor{famGemma}{HTML}{762a83}
\definecolor{famJanus}{HTML}{d6604d}

\newcommand{\mQwen}[1]{\textcolor{famQwen}{\textbf{#1}}}
\newcommand{\mLLaVA}[1]{\textcolor{famLLaVA}{\textbf{#1}}}
\newcommand{\mIntern}[1]{\textcolor{famIntern}{\textbf{#1}}}
\newcommand{\mGemma}[1]{\textcolor{famGemma}{\textbf{#1}}}
\newcommand{\mJanus}[1]{\textcolor{famJanus}{\textbf{#1}}}

\definecolor{statGated}{HTML}{b2182b}
\definecolor{statOpen}{HTML}{2e7d32}
\definecolor{statPartial}{HTML}{e08214}

\newcommand{\statGated}[1]{\textcolor{statGated}{\textbf{#1}}}
\newcommand{\statOpen}[1]{\textcolor{statOpen}{\textbf{#1}}}
\newcommand{\statPartial}[1]{\textcolor{statPartial}{\textbf{#1}}}

\definecolor{tintQwen}{HTML}{eaf2f8}
\definecolor{tintLLaVA}{HTML}{fcf2e6}
\definecolor{tintIntern}{HTML}{ecf6ee}
\definecolor{tintGemma}{HTML}{f3ecf6}
\definecolor{tintJanus}{HTML}{fbece9}

\graphicspath{{figures/}}

\title{Knowing Isn't Always Saying: \\
  When Do Spatial Encodings Reach Answers in Vision-Language Models?}

\author{
  Zeyu Wang\textsuperscript{*} \\
  Peking University \\
  Beijing, China \\
  \texttt{2501210409@stu.pku.edu.cn} \\
  \And
  Xinming Xu\textsuperscript{*} \\
  Tsinghua University \\
  Beijing, China \\
  \texttt{xu-zh23@mails.tsinghua.edu.cn} \\
}

\begin{document}
\maketitle
\begingroup
\renewcommand{\thefootnote}{*}
\footnotetext{Equal contribution.}
\endgroup

\begin{abstract}
Vision-language models are known to encode spatial information in their
hidden states, yet often fail to use it when answering. However, it
remains unclear when and where this encoded information reaches the
answer. We address this with direction patching, a class-conditioned causal
intervention applied across layers, token positions, and prompt formats. Using spatial-ID directions constructed following prior encoding evidence,
we find that causal influence on answer logits emerges only at mid-to-deep
depths. Text chain-of-thought suppresses immediate object-word
argmax-level transport in most models, while visually grounded prompts
keep it open. Positive target-logit gain can remain below the argmax
threshold, and transport can re-emerge at the final prefix token or at
the answer step in deeper layers. Across the ten VLMs we study, these local effects form descriptive transport patterns. Complementary experiments characterize how these patterns shift across datasets, attributes, and encoding amplitudes. Together, these results reframe the encoding-grounding gap as a problem of conditional transport in VLMs.

\end{abstract}

\section{Introduction}
\label{sec:intro}

Vision-Language Models have achieved remarkable progress across a wide range of visual reasoning tasks.
However, recent interpretability studies reveal a puzzling dissociation between representation and behavior:
models can correctly encode visual attributes in their hidden states,
yet often fail to use that information when producing answers
\citep{nooralahzadeh2026arbitration, liu2026seeingnotbelieving, asadi2026mirage}.
Chain-of-Thought prompting is commonly adopted to mitigate reasoning failures,
but on spatial tasks it can produce the opposite effect:
introducing text CoT often substantially degrades performance across multiple open-source VLMs.

\begin{figure*}[t]
\centering
\includegraphics[width=0.95\textwidth]{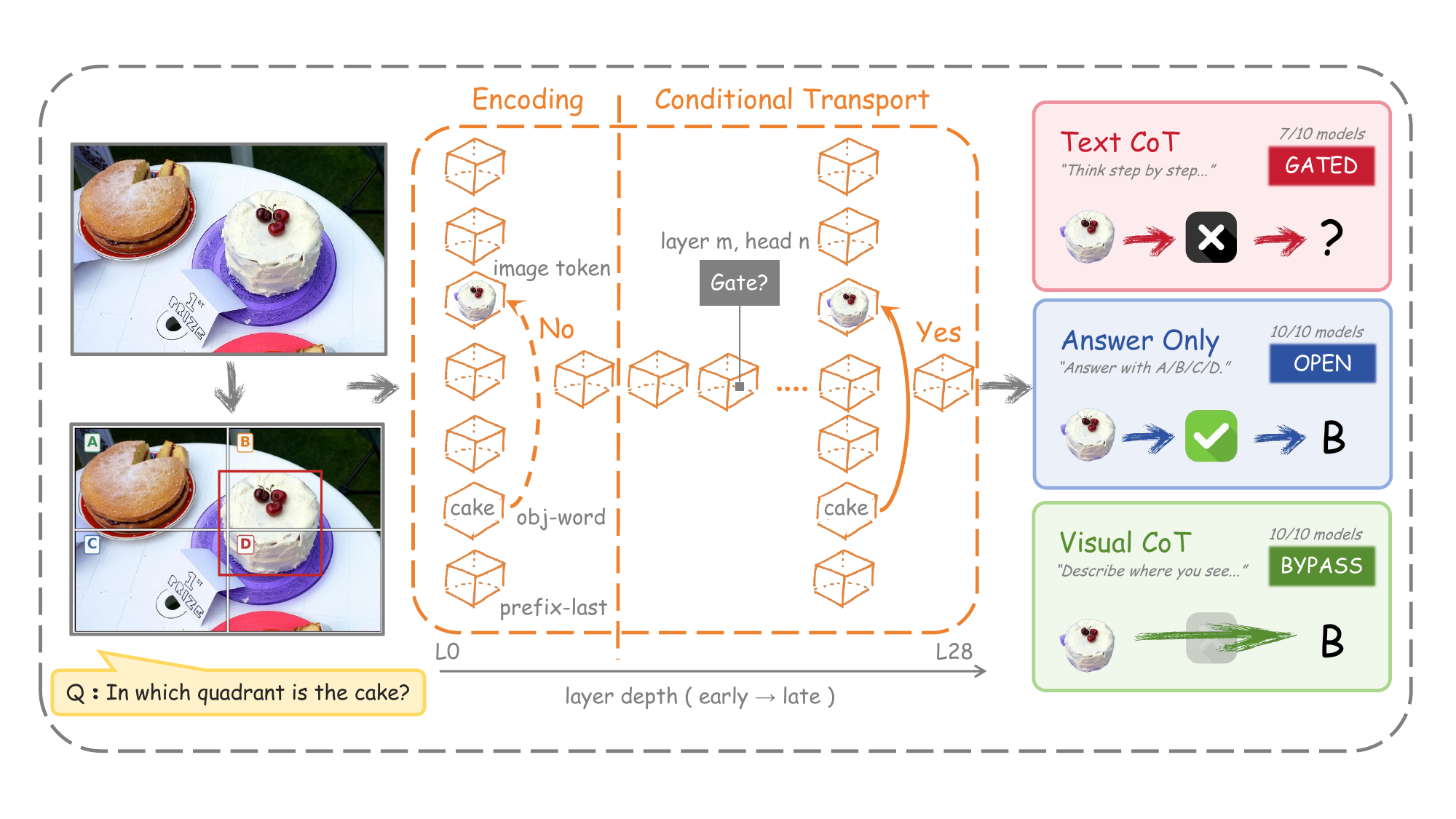}
\caption{Direction patching reveals conditional transport.
Spatial encoding is broadly present across VLMs, but whether it reaches the answer depends on prompt type, token position, and layer depth.
Text CoT can block object-word transport, whereas visual grounding prompts keep this pathway open.}
\label{fig:teaser}
\end{figure*}

To understand this paradox, we move beyond static representation analysis and instead trace the causal dynamics of information flow.
Prior work has established the encoding-utilization dissociation
\citep{li2025groundingme, kamath2023whatsup},
but it remains unclear under what local conditions encoded visual information causally reaches the answer.
The missing piece is a transport-level account:
when the spatial signal becomes causally effective,
where it is routed inside the prompt,
and how reasoning prompts change that route.

We investigate these questions using \emph{direction patching}
\citep{kang2026linearmechanisms},
a causal intervention inspired by activation patching
\citep{vig2020causalmediation, wang2023ioi}.
The method manipulates class-conditioned hidden-state directions at controlled layers and token positions,
isolating their direct causal effect on answer logits.
We build on the spatial-ID directions established by \citet{kang2026linearmechanisms}
and contribute the first systematic mapping of how these directions transport
to answer logits across layer, position, and prompt format.

Applying direction patching to ten VLMs across three datasets and seven prompt formats,
we analyze spatial transport at both a local and a model-level scale.
Three local regularities emerge.
\textit{Late emergence}: the spatial-ID direction is constructed from
early-layer representation evidence, but causal transport activates only
at L12--L32.
\textit{CoT gating}: text CoT suppresses immediate object-word
argmax-level transport in most models, whereas visual grounding prompts
keep it open.
\textit{Position dependence}: spatial transport can reappear at the
final prefix token in deeper layers, making the pathway jointly governed
by prompt format and token position.

At the model level, these local patterns organize into a small number of
descriptive transport groupings rather than a single uniform behavior.
We further test the logit-level map with generate-mode behavioral experiments,
where steering interventions induce substantial accuracy changes under
open transport pathways,
and with a head-knockout probe that provides initial, architecture-specific
causal evidence for a CoT-specific gating component in one architecture.
The patterns also replicate across datasets and expose attribute- and amplitude-dependent boundaries.
Together, the results provide a mechanistic mapping of how text CoT can
alter spatial transport
\citep{kancheti2026cotdegrades}
and identify visual grounding prompts
\citep{jiang2025vlmr3, qin2025chain}
as a prompt-only intervention that keeps the object-word pathway open.

\section{Experiment}
\label{sec:setup}

\subsection{Tasks}

We evaluate on three four-way MCQ tasks that share a common pipeline but
target different visual attributes.
\textbf{RefCOCO spatial}: from RefCOCO/RefCOCO+ \citep{refcoco2014,refcoco_yu2016}
referring expressions, each instance becomes \textit{``In which quadrant
is the [object]? (A) top-left (B) top-right (C) bottom-left (D)
bottom-right''}, with the gold label set by the bounding-box centroid.
Ambiguous boundary cases are filtered.
We use $n{=}50$ per quadrant for 7B/8B models and $n{=}30$ for Qwen-32B,
Gemma-3, and Janus.
\textbf{GQA spatial} uses analogous four-quadrant questions derived from
GQA scene graphs \citep{gqa2019}, providing different images and object
distributions.
\textbf{GQA color} replaces the spatial question with \textit{``What
color is the [object]?''} (red / blue / white / black);
the centroid computation, crossfit, and $\alpha$ settings are identical,
only the patched direction changes from spatial-ID to color-ID.

\subsection{Models}

\begin{table}[t]
\centering
\small
\setlength{\tabcolsep}{3pt}
\renewcommand{\arraystretch}{1.05}
\begin{tabular}{l l r r}
\toprule
Model & Family & Params & Layers \\
\midrule
\rowcolor{tintQwen}   \mQwen{Qwen2.5-VL-7B}        & Qwen     & 7B  & 28 \\
\rowcolor{tintQwen}   \mQwen{Qwen2.5-VL-32B}       & Qwen     & 32B & 64 \\
\rowcolor{tintLLaVA}  \mLLaVA{LLaVA-OneVision-7B}  & LLaVA    & 7B  & 28 \\
\rowcolor{tintIntern} \mIntern{InternVL2.5-8B}     & InternVL & 8B  & 32 \\
\rowcolor{tintIntern} \mIntern{InternVL3-8B}       & InternVL & 8B  & 28 \\
\rowcolor{tintIntern} \mIntern{InternVL3-14B}      & InternVL & 14B & 48 \\
\rowcolor{tintGemma}  \mGemma{Gemma-3-4B-IT}       & Gemma    & 4B  & 34 \\
\rowcolor{tintGemma}  \mGemma{Gemma-3-12B-IT}      & Gemma    & 12B & 48 \\
\rowcolor{tintGemma}  \mGemma{Gemma-3-27B-IT}      & Gemma    & 27B & 62 \\
\rowcolor{tintJanus}  \mJanus{Janus-Pro-7B}        & DeepSeek & 7B  & 30 \\
\bottomrule
\end{tabular}
\caption{Ten evaluated VLMs spanning five families,
including three within-family scale pairs:
Qwen 7B/32B, InternVL3 8B/14B, and Gemma-3 4B/12B/27B.
Row tints group rows by family.}
\label{tab:models}
\end{table}

\noindent
We evaluate ten open-weight VLMs from five families (Table~\ref{tab:models}):
Qwen2.5-VL-7B and -32B \citep{qwen25vl2025},
LLaVA-OneVision-7B \citep{llavaonevision2024},
InternVL2.5-8B \citep{internvl25_2024},
InternVL3-8B and -14B \citep{internvl3_2025},
Gemma-3-4B, -12B, and -27B \citep{gemma3_2025},
and DeepSeek-Janus-Pro-7B \citep{janus2025}.
All models expose residual-stream activations required for direction patching.
The set includes three within-family scale pairs
(Qwen 7B/32B, InternVL3 8B/14B, Gemma-3 4B/12B/27B),
enabling controlled comparisons of how model scale affects transport.

\subsection{Method: Direction Patching}
\label{sec:setup:method}

We build on the spatial-ID encoding reported by \citet{kang2026linearmechanisms}.
At each layer $\ell$, we compute class-conditioned centroids $C^{(\ell)}_q$
(one per quadrant or color) from hidden states at the object-word token position,
using a five-fold crossfit split to avoid circularity.
The intervention shifts the residual-stream activation at position $p$, layer $\ell$:
\[
  h^{(\ell)}_p \;\leftarrow\; h^{(\ell)}_p
    + \alpha \bigl(C^{(\ell)}_{\text{target}} - C^{(\ell)}_{\text{source}}\bigr),
\]
where $\alpha{=}5$ for obj\_word and $\alpha{=}10$ for prefix\_last unless stated otherwise.
These canonical values were chosen from an $\alpha \in \{1,2,5,10\}$ sweep
as the midpoint of the monotonically increasing dose-response range
(App.~\ref{sec:appendix:alpha}).
The model then completes a forward pass and we read next-token logits over \{A, B, C, D\}.

\paragraph{Metric.}
$\Delta$argmax $=$ target flip rate under intervention $-$ target flip rate under a same-norm random-direction baseline.
A positive value indicates that the patched direction causally shifts the model's answer toward the target class
beyond what a generic perturbation achieves. We also report the continuous
target-logit gain, defined as the mean target-letter logit under the
direction patch minus the corresponding mean under the random-direction
baseline. Thus, target-logit gain measures directional influence even when
the four-way argmax does not change. We use generate-mode experiments
(128 tokens) to measure the change in parsed-answer accuracy ($\Delta$acc)
as a separate behavioral validation. For these runs, we additionally
report $\Delta_{\mathrm{parse}}$, the intervention-minus-noise contrast in
the final parsed target rate; it is distinct from both $\Delta$acc and the
fixed-state next-token $\Delta$argmax.

\paragraph{Operational definition of transport.}
For the coarse answer-option diagnostic, we use \emph{argmax-level transport}
to mean a non-zero $\Delta$argmax under direction patching at a given layer
$\ell$, token position $p$, and prompt format. A zero $\Delta$argmax does
not imply zero target-logit gain. The measurement is causal evidence that
the spatial direction influences the answer competition along the
linear subspace identified by \citet{kang2026linearmechanisms}; it is silent
on transport that may occur through non-linear pathways or alternative
directions outside this subspace (see Limitations).

\paragraph{Intervention positions.}
We patch at two positions:
\textbf{obj\_word}, the token span matched to the mentioned object
expression (falling back to content-word anchor tokens when an exact span
match is unavailable),
and \textbf{prefix\_last}, the final token of the input sequence before generation begins.
The prefix\_last experiments intentionally reuse the object-word-derived
spatial direction rather than recomputing a position-specific centroid;
they therefore test whether the same direction can be read out at a later
position.

\paragraph{Prompt formats.}
We use seven instruction suffixes appended to a shared question stem,
spanning a reasoning-length spectrum from \emph{cot}
(``Let's think step by step'') through \emph{brief\_reason},
\emph{answer\_first}, \emph{final\_tag}, and \emph{answer\_only}
(``Answer with only the letter''), plus two visual-grounding prompts
(\emph{visual\_cot}, \emph{visual\_direct}) that direct reasoning toward
visual observation rather than abstract spatial logic.
Verbatim templates are in App.~\ref{sec:appendix:prompts}.


\section{Local Dynamics of Conditional Transport}
\label{sec:transport}

We first examine conditional transport at the local scale, across controlled layers and token positions.
Spatial-ID directions can be available before they causally affect the answer (\S\ref{sec:transport:late}).
Transport from encoding to answer is then conditional along two axes: prompt type (\S\ref{sec:transport:cot}, \S\ref{sec:transport:visual}) and token position (\S\ref{sec:transport:position}).
Layer depth tells us \emph{when} the pathway activates; token position tells us \emph{where} it is routed.

\begin{figure*}[!t]
\centering
\includegraphics[width=0.90\textwidth]{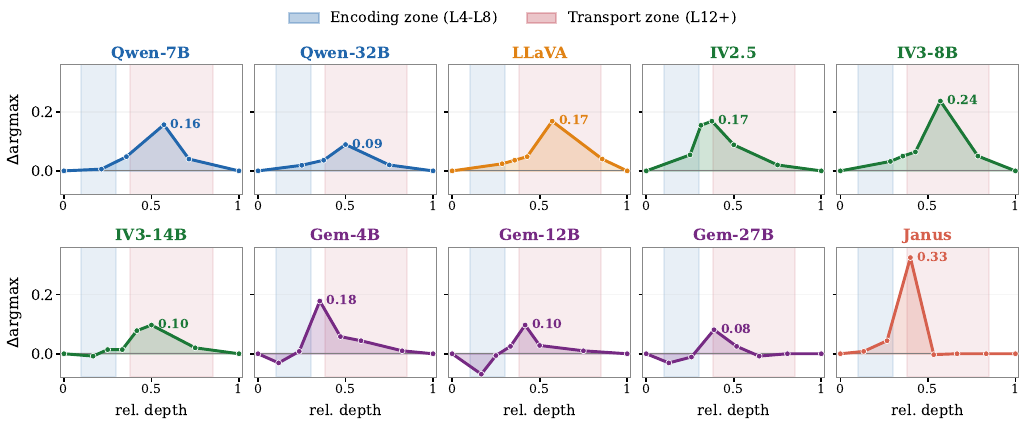}
\caption{Layer-wise $\Delta$argmax for ten models (obj\_word, answer\_only, $\alpha{=}5$).
Argmax-level transport activates only at mid-to-deep layers, although the
Kang-style spatial-ID direction is constructed from early-layer
representation evidence.}
\label{fig:late_emergence}
\end{figure*}

\subsection{Spatial Encoding Does Not Immediately Become Transport}
\label{sec:transport:late}
\label{sec:encoding}

Following \citet{kang2026linearmechanisms},
we extract object-word hidden states and compute per-quadrant centroids via five-fold crossfit.
We take the early-layer class-contrastability used to construct this
direction as prior encoding evidence; this subsection does not introduce
an independent separability or probing curve.
Under the answer-only prompt, direction patching at the object-word position yields
positive $\Delta$argmax in all ten models,
with peak transport clustering in L8--L16 for 7B/8B models
and shifting to L20--L32 for larger models.
The same object-word centroid construction is used across prompt formats,
so the prompt effects reported below are not attributable to changing the patched spatial direction.
The question is whether these available spatial directions influence the answer
as soon as they are detectable, or only after later-layer processing.

Figure~\ref{fig:late_emergence} plots $\Delta$argmax across layers for all ten models under the answer-only prompt at the object-word position.
Shallow layers (L0--L8) produce near-zero $\Delta$argmax in every model,
even though the spatial-ID direction is available from the early-layer
representation used for its construction.
Transport activates in mid-to-deep layers
(Qwen-7B L16 $+0.156$; Gemma-3-4B L12 $+0.178$;
Janus L12 $+0.325$, the largest peak we observe).
For larger models the peak appears at a higher absolute layer index
(Qwen-32B L32 of 64, $+0.089$), although the depth relative to total
layer count is comparable to the 7B/8B models.

The gap between early-layer direction availability (L4--L8) and transport
depth (L12--L32) shows that representation and causal use are dissociable
within a single forward pass.
Object-word tokens carry spatial directions from early layers,
but those directions do not causally shift the answer distribution until many layers later.
Thus, detecting a spatial code is not sufficient to show that the code
is being transmitted to the answer; transport has its own layer condition.

\paragraph{Controls.}
Wrong-axis and non-object-position patching both stay within an
order of magnitude of zero ($[-0.018,\allowbreak +0.031]$ and
$[-0.017,\allowbreak +0.006]$, respectively) across all ten models,
confirming the effect is direction- and position-specific
(App.~\ref{sec:appendix:controls}).

\subsection{CoT Suppresses Immediate Object-Word Transport}
\label{sec:transport:cot}
Having established an answer-only transport pathway at obj\_word,
we next change only the prompt: from answer\_only to standard chain-of-thought
(``Let's think step by step'').

At the object-word position, this single change suppresses the immediate
argmax-level transport established in the previous subsection.
Eight of ten models show CoT peak $\Delta$argmax $\leq 0.006$ at obj\_word
(we use $<0.01$ as a descriptive near-zero threshold; six of these are
exactly zero and LLaVA and Qwen-32B are borderline near-zero cases),
including the model with the strongest answer-only peak: Janus drops from $+0.325$ to below this resolution.
The three Gemma-3 variants are uniformly near-zero regardless of scale,
and Qwen-32B sits within the noise floor at $+0.003$.
The two exceptions are Qwen-7B ($+0.120$) and InternVL2.5 ($+0.042$, $75\%$ reduction from answer-only);
both are analysed in \S\ref{sec:variation}.
Per-model peak values and bootstrap CIs are in App.~\ref{sec:appendix:peak_values}.

\paragraph{Gating is argmax-level, not signal erasure.}
If CoT merely weakened the signal, increasing $\alpha$ should systematically
lift CoT $\Delta$argmax above the near-zero range.
Sweeping $\alpha \in \{1,2,5,10\}$ (App.~\ref{sec:appendix:alpha}),
answer-only grows across the dose range, whereas the suppressed models
remain at or near zero; the two borderline cases do not show a systematic
rise into the open range.
The block is at the argmax-flip level, not at the underlying logit:
under CoT, IV3-8B's $\Delta$argmax is $0$ while the target-letter
logit gain remains $+0.421$ (App.~\ref{sec:appendix:ci}), a sub-threshold
directional signal the head knockout in
\S\ref{sec:variation:validation} exploits.
This connects to the accuracy-side observation of
\citet{kancheti2026cotdegrades}, who report CoT-induced spatial accuracy
degradation without identifying a mechanism:
in our data, CoT suppresses the obj\_word argmax-level route in the
majority, so the patched direction does not reliably change the answer
option competition at that position.

\paragraph{Answer-step timing control.}
The canonical readout fixes the prompt state, layer, and token position so
that it measures a comparable immediate answer-option effect. To test
whether this choice misses a later answer readout, we generated a clean
CoT prefix, located the model's actual answer-letter step, reconstructed the
prefix immediately before that letter, and repeated the same obj\_word
direction patch. In LLaVA-OV-7B, Qwen2.5-VL-7B, and InternVL3-8B, the
target-logit gain is positive across L8--L20 and fades at L24; at L16 the
values are $+0.430$, $+0.309$, and $+0.538$, respectively
(App.~\ref{sec:appendix:answer_step}). This timing control supports the
interpretation that the first-step result is not caused only by a mismatch
between the fixed readout and the eventual answer step. It does not imply
that the effect is identical at every generated token.

\subsection{Visual Grounding Bypasses the Gate}
\label{sec:transport:visual}

\begin{table}[t]
\centering
\footnotesize
\setlength{\tabcolsep}{3pt}
\renewcommand{\arraystretch}{0.98}
\begin{tabular}{l rrr}
\toprule
Model & CoT & Vis-CoT & Ans-only \\
\midrule
\rowcolor{tintQwen}   \mQwen{Qwen-7B}       & +.081 & +.136 & +.169 \\
\rowcolor{tintQwen}   \mQwen{Qwen-32B}      & +.006 & \statOpen{+.117}$^{\ast}$ & +.089 \\
\rowcolor{tintLLaVA}  \mLLaVA{LLaVA}        & .000  & +.103 & +.158 \\
\rowcolor{tintIntern} \mIntern{InternVL2.5} & .000  & \statOpen{+.153}$^{\ast}$ & +.114 \\
\rowcolor{tintIntern} \mIntern{InternVL3-8B}  & .000  & +.164 & +.236 \\
\rowcolor{tintIntern} \mIntern{InternVL3-14B} & .000  & \statOpen{+.117}$^{\ast}$ & +.097 \\
\rowcolor{tintGemma}  \mGemma{Gemma-3-4B}   & .000  & +.092 & +.178 \\
\rowcolor{tintGemma}  \mGemma{Gemma-3-12B}  & .000  & +.064 & +.097 \\
\rowcolor{tintGemma}  \mGemma{Gemma-3-27B}  & .000  & +.022 & +.081 \\
\rowcolor{tintJanus}  \mJanus{Janus}        & .000  & \statOpen{+.403}$^{\ast}$ & +.069 \\
\bottomrule
\end{tabular}
\caption{$\Delta$argmax at obj\_word at each model's visual\_cot peak layer (RefCOCO, $\alpha{=}5$).
$^{\ast}$: visual\_cot exceeds answer\_only.
All ten models recover positive transport under visual\_cot, including eight suppressed under text CoT.}
\label{tab:visual_bypass}
\end{table}

The previous subsection established that text CoT suppresses immediate spatial transport at obj\_word.
Is the gating triggered by having intermediate reasoning tokens, or specifically by text-abstract reasoning? We keep the intervention fixed at obj\_word and vary the prompt attribute.

We compare text CoT to two visual-grounding prompts that also generate
intermediate tokens but redirect the chain toward visual evidence
(\emph{visual\_cot}: ``describe where you see the object'';
\emph{visual\_direct}: ``note its position relative to the center'';
full templates in App.~\ref{sec:appendix:prompts}).

Table~\ref{tab:visual_bypass} compares peak $\Delta$argmax at obj\_word
for text CoT, visual\_cot, and answer-only.
All ten models recover positive obj\_word $\Delta$argmax under
visual\_cot, including the eight that are suppressed under text CoT.
Four models produce visual\_cot effects that even exceed answer-only:
Qwen-32B ($+.117$ vs.\ $+.089$), IV2.5 ($+.153$ vs.\ $+.114$),
IV3-14B ($+.117$ vs.\ $+.097$), and most strikingly Janus
($+.403$ vs.\ $+.069$, a $5.8\times$ amplification).
The Janus jump shows visual grounding does not just ``unblock'' but
actively elicits a spatial direction that answer-only under-extracts.
Because visual\_cot generates intermediate tokens like text CoT, a
token-count explanation for obj\_word gating fails: the conditional
variable is the reasoning format
(text-abstract ``think step by step'' closes the pathway;
visual grounding ``describe what you see'' leaves it open).

\subsection{Transport Pathways Are Position-Dependent}
\label{sec:transport:position}

\begin{figure*}[!t]
\centering
\includegraphics[width=0.88\textwidth]{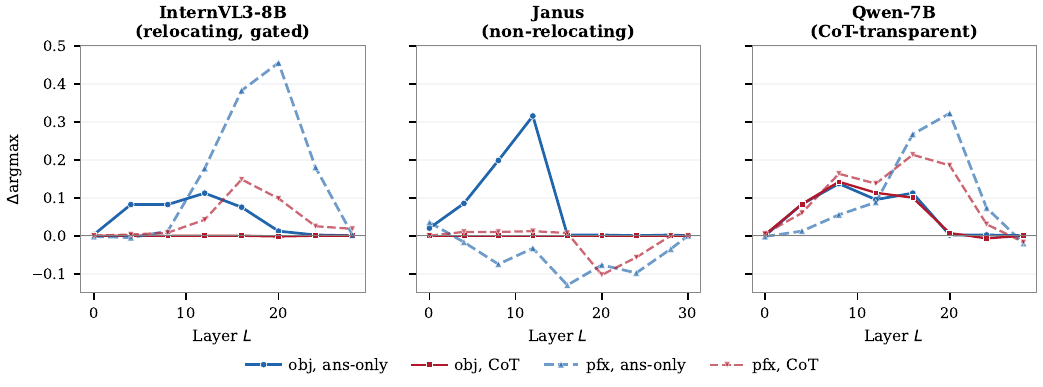}
\caption{Position-dependent transport (GQA Spatial, three models).
Left: InternVL3-8B shows the relocation pattern (obj\_word L12 $\to$ prefix\_last L20; CoT gated only at obj\_word).
Center: Janus concentrates at obj\_word. Right: Qwen-7B transports at both positions under both prompts.}
\label{fig:position}
\end{figure*}

Patching at prefix\_last (the final token before generation) tests
whether closing the obj\_word pathway removes spatial transport entirely
or leaves a later-position route open.
Here we reuse the object-word-derived spatial direction and change only
the injection position, so the comparison tests relocation/readout
compatibility rather than a separately optimized prefix\_last basis.
We use GQA Spatial for its full position $\times$ prompt grid at
$\alpha{=}5$; the RefCOCO prefix\_last ladder
(App.~\ref{sec:appendix:ladder}) gives converging evidence at $\alpha{=}10$.

Most models show a relocation pattern (Figure~\ref{fig:position}):
obj\_word $\Delta$argmax peaks in early-to-mid layers and declines,
while prefix\_last $\Delta$argmax emerges deeper.
InternVL3-8B is the clearest case (obj\_word L12 $+0.112$, prefix\_last
L20 $+0.455$), and CoT is near-zero at obj\_word but \emph{not} at prefix\_last.
The suppression is therefore local to one transport route, and the conditional
pathway is jointly determined by prompt type and token position.

We call this two-step shape (obj\_word peak in early-to-mid layers,
prefix\_last re-emergence in deeper layers) the \emph{relocation pattern}
and use this term throughout \S\ref{sec:variation}.
Not all models follow it: Janus concentrates transport at obj\_word
($+0.315$ at L12) with prefix\_last $\Delta$argmax going negative
(patching pushes the answer \emph{away} from target, indicating that
prefix\_last is dominated by competing features rather than spatial
content), and Gemma-3 shows only weak, dataset-dependent prefix\_last
effects. These differences motivate \S\ref{sec:variation:patterns}.

\section{Model-Level Patterns and Boundary Conditions}
\label{sec:variation}

\S\ref{sec:transport} isolated the local conditions under which spatial information reaches the answer:
layer depth, prompt type, and token position.
We now move to a model-level view:
whether these local patterns organize into structured cross-model variation
(\S\ref{sec:variation:patterns}),
whether the logit-level map is reflected beyond next-token logits
(\S\ref{sec:variation:validation}),
and where the map transfers or changes across datasets, attributes, and encoding amplitudes
(\S\ref{sec:variation:boundaries}).

\subsection{Transport Groupings Across Models}
\label{sec:variation:patterns}

\begin{figure*}[t]
\centering
\includegraphics[width=0.90\textwidth]{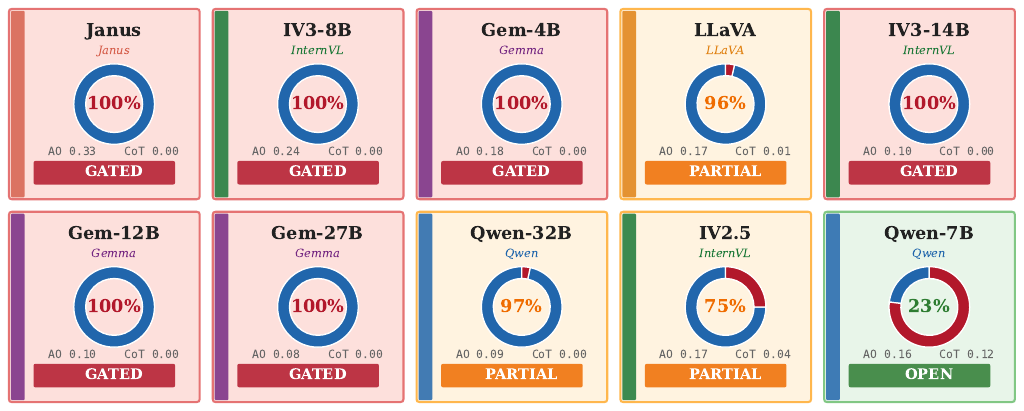}
\caption{Per-model CoT gating signature (obj\_word, RefCOCO, $\alpha{=}5$).
Eight of ten models fall below the descriptive $0.01$ threshold:
six are exact-zero and two are near-zero/borderline; Qwen-7B retains
substantial CoT transport.
Numerical values in App.~\ref{sec:appendix:peak_values}.}
\label{fig:cot_gating}
\end{figure*}

\begin{table}[t]
\centering
\small
\setlength{\tabcolsep}{3pt}
\renewcommand{\arraystretch}{1.15}
\begin{tabular}{p{1.6cm}p{2.8cm}cc}
\toprule
Pattern & Models & \makecell{Obj.\\route} & Reloc. \\
\midrule
Open              & \mQwen{Qwen-7B}
                  & \statOpen{Open}  & \statOpen{Yes} \\
\addlinespace
\makecell[l]{Reduced\\supp.,reloc.}
                  & \makecell[l]{\mLLaVA{LLaVA}, \mIntern{IV2.5},\\
                    \mIntern{IV3-8B}, \mIntern{IV3-14B}}
                  & \statPartial{Supp./red.} & \statOpen{Yes} \\
\addlinespace
\makecell[l]{Suppressed,\\non-reloc.}
                  & \makecell[l]{\mGemma{Gemma-3}\\(4/12/27B),\\\mJanus{Janus}}
                  & \statGated{Supp.} & \statGated{No} \\
\addlinespace
\makecell[l]{Prompt-\\selective}  & \mQwen{Qwen-32B}
                  & \statPartial{Near-zero} & \statPartial{Sel.} \\
\bottomrule
\end{tabular}
\caption{Four descriptive transport patterns across ten models.
``Reloc.'': whether transport reappears at prefix\_last (Yes / Sel.\ = prompt-selective / No).
Colors match Figure~\ref{fig:cot_gating}.}
\label{tab:signatures}
\vspace{-5.5mm}
\end{table}

Figure~\ref{fig:cot_gating} summarizes the per-model gating signatures,
and Table~\ref{tab:signatures} organizes the ten models into four
descriptive patterns.
The \emph{reduced/suppressed, relocating} pattern is the modal pattern
(4/10 models): LLaVA and the three InternVL variants suppress or sharply
reduce transport at obj\_word but show positive prefix\_last transport in
deeper layers.
The \emph{suppressed, non-relocating} pattern (4/10) comprises the three
Gemma-3 variants and Janus: spatial information either stays at the
object-word token (Janus, $+0.325$ at L12) or does not recover at
prefix\_last (Gemma-3).
Qwen-7B is the sole \emph{open} model: CoT does not block transport
at either position; \S\ref{sec:variation:boundaries} treats this as an
amplitude-sensitive boundary case.
Qwen-32B is \emph{prompt-selective}: near-zero CoT transport at obj\_word
but a prompt-format-dependent prefix\_last pathway (final\_tag $+0.431$
vs.\ answer-only $+0.019$ at L55; App.~\ref{sec:appendix:ladder}).
With $N{=}10$ models, two of the four patterns have a single member,
so these labels summarize the observed combinations rather than define a
statistically validated classification.
Grouping is consistent within family in two of three multi-scale families
(InternVL3-8B and -14B share identical gating; Gemma-3 4B/12B/27B all
fall in the suppressed, non-relocating pattern), but the Qwen family
shifts from open (7B) to prompt-selective (32B), so family is not a
sufficient predictor of grouping in our data.

\noindent\textbf{Janus: dual role.}
Janus is the strongest visual-bypass case
(\S\ref{sec:transport:visual}): visual\_cot lifts obj\_word $\Delta$argmax
from $+0.069$ to $+0.403$, a $5.8\times$ amplification no other model approaches.
The same model is also the cleanest counter-example to a strong reading
of ``linear-subspace transport $\Rightarrow$ behaviour'':
despite the $+0.403$ logit effect, generate-mode steering changes
accuracy by $0$pp (baseline $50$--$58\%$), suggesting that Janus's
unified architecture routes generation through a pathway misaligned with
our linear subspace.
This is a concrete boundary for the operational definition in
\S\ref{sec:setup:method} (see Limitations).

\subsection{Behavioral Consequences and Head-Level Probe}
\label{sec:variation:validation}

We next test whether the logit-level map has support beyond next-token
logits.

\paragraph{Behavioral validation.}
Full 128-token generation with the same intervention provides a behavioral
check on the logit-level map.
Under answer-only, steering toward incorrect quadrants drops accuracy
by $8$--$21$pp in nine of ten models
(Qwen-7B $74.2\%\!\to\!53.6\%$; IV3-8B $77.5\%\!\to\!56.9\%$;
App.~\ref{sec:appendix:accuracy}).
Janus is the boundary case, with no answer-only accuracy drop in this
generation protocol.
Under CoT, models with intact CoT baselines (LLaVA $55.8\%$,
IV3-14B $60.8\%$, Gem-12B $45.8\%$) still show $13$--$16$pp accuracy
drops under the same object-word intervention, even though their
fixed-state obj\_word $\Delta$argmax is near zero for CoT.
Because the intervention is applied before autoregressive generation,
the perturbation can affect later hidden states and the final parsed
answer without changing the immediate answer-option argmax. The
independent prefix\_last sweep identifies a compatible later readout
route in relocating models; the generate-mode table itself does not
patch prefix\_last.
Under visual\_cot, steering produces sizable drops for many models,
while IV2.5, Gemma-3, and Janus show weaker changes. This pattern is
consistent with visual grounding preserving an active spatial pathway,
but also highlights the model dependence of generate-mode validation.

For Qwen-32B and IV2.5, CoT collapses the baseline well below 4-way
chance ($9.2\%$, $4.2\%$); the $\leq 2.2$pp post-steering drop is
consistent with logit-level gating but cannot be cleanly separated from
generic CoT derailing, so behavioral evidence rests on the
intact-baseline group above.

\begin{figure*}[t]
\centering
\includegraphics[width=0.78\textwidth]{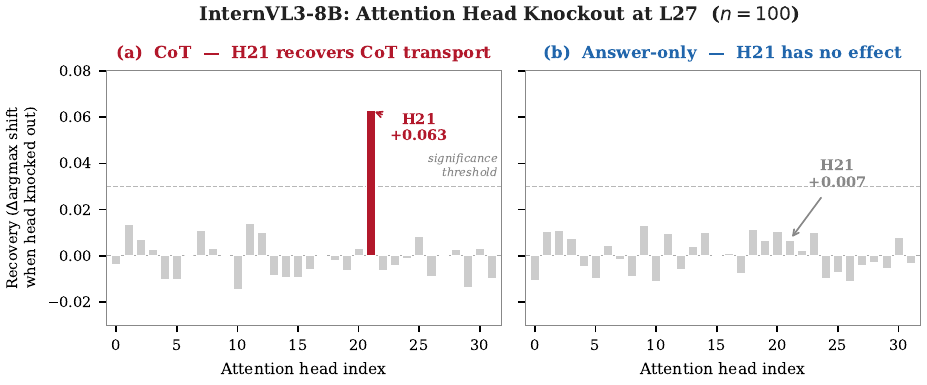}
\caption{Attention head knockout at L27 of InternVL3-8B ($n{=}100$, RefCOCO, $\alpha{=}5$).
(a)~Under CoT, knocking out H21 recovers $+0.063$ $\Delta$argmax, the only head above the significance threshold among all 32 heads.
(b)~Under answer-only, H21 has no effect ($+0.007$).
H21 is a CoT-specific gating candidate in this scan, rather than a general transport component.}
\label{fig:knockout}
\end{figure*}

\paragraph{Head-level evidence: proof of concept.}
We sweep attention head knockouts downstream of the steer layer:
for each of 32 heads at a candidate layer, we zero its output projection
and measure CoT $\Delta$argmax recovery at obj\_word.
In InternVL3-8B, a single head L27.H21 produces $+0.063$ recovery under
CoT ($n{=}100$; CoT baseline $+0.095$), while all 31 other heads stay
within $\pm 0.02$ of baseline (Figure~\ref{fig:knockout}a).
H21 has no effect under answer-only (Figure~\ref{fig:knockout}b,
$+0.007$), consistent with a CoT-specific gating role.
A parallel scan in LLaVA-OV-7B finds no comparable concentration:
the best candidate (L22.H25) recovers only $+0.013$ at $n{=}100$
(App.~\ref{sec:appendix:knockout}).
We do not test Qwen-7B because it is open at obj\_word and has no
suppressed route to localize.
Head-level concentration is thus evident in at least one architecture
but architecture-dependent;
a full circuit account (upstream triggers, downstream completion,
cross-architecture comparison) remains open.

\subsection{Generality and Boundary Conditions}
\label{sec:variation:boundaries}

\begin{table}[t]
\centering
\small
\setlength{\tabcolsep}{4pt}
\renewcommand{\arraystretch}{1.15}
\begin{tabular}{p{3.2cm} c p{2.7cm}}
\toprule
Prediction & Match & Note \\
\midrule
CoT suppression at obj\_word    & \statOpen{\textbf{9/10}}
                                & \mQwen{Qwen-7B}: remains open \\
Peak layer ${\pm}$8L            & \statOpen{\textbf{6/6}}
                                & original 6 models \\
pfx\_last deeper than obj\_word & \statOpen{\textbf{8/10}}
                                & \mJanus{Janus}, \mGemma{Gem-27B} anomalous \\
\bottomrule
\end{tabular}
\caption{Structural transfer from RefCOCO to GQA Spatial.
Three transport-map predictions tested with independent images, scene graphs, and object distributions.}
\label{tab:cross_dataset}
\end{table}

We test the generality of the transport map across datasets, visual
attributes, and encoding amplitudes.

\paragraph{Cross-dataset generalization.}
We derive four-quadrant spatial MCQs from GQA scene graphs \citep{gqa2019}
and test three structural predictions from the RefCOCO map
(Table~\ref{tab:cross_dataset}):
CoT suppression status transfers in 9/10 models, peak layers stay within $\pm 8$
layers in the 6 models with full coverage, and 8/10 reproduce the
relocation pattern.
Janus stays non-relocating; Gemma-3-27B shows a weak prefix\_last effect
($+0.187$ at L40) not captured by the cross-dataset ladder.

\paragraph{Cross-attribute generalization.}
\begin{table}[t]
\centering
\footnotesize
\setlength{\tabcolsep}{3pt}
\renewcommand{\arraystretch}{1.02}
\begin{tabular}{l rr r}
\toprule
Model & Spatial peak & Color peak & Shift \\
\midrule
\rowcolor{tintQwen}   \mQwen{Qwen-7B}        & L8  & L4  & \statOpen{$-4$} \\
\rowcolor{tintQwen}   \mQwen{Qwen-32B}       & L24 & L16 & \statGated{$\mathbf{-8}$} \\
\rowcolor{tintLLaVA}  \mLLaVA{LLaVA}         & L8  & L4  & \statOpen{$-4$} \\
\rowcolor{tintIntern} \mIntern{InternVL2.5}  & L8  & L4  & \statOpen{$-4$} \\
\rowcolor{tintIntern} \mIntern{InternVL3-8B} & L12 & L4  & \statGated{$\mathbf{-8}$} \\
\rowcolor{tintIntern} \mIntern{InternVL3-14B}& L12 & L12 & \statPartial{$0$} \\
\rowcolor{tintGemma}  \mGemma{Gemma-3-4B}    & L8--12 & L8 & \statPartial{$0$ to $-4$} \\
\rowcolor{tintGemma}  \mGemma{Gemma-3-12B}   & L18 & L12 & \statGated{$-6$} \\
\rowcolor{tintGemma}  \mGemma{Gemma-3-27B}   & L24 & L16 & \statGated{$\mathbf{-8}$} \\
\rowcolor{tintJanus}  \mJanus{Janus}         & L12 & L8  & \statOpen{$-4$} \\
\bottomrule
\end{tabular}
\caption{Obj\_word peak layer on GQA Spatial vs.\ GQA Color (answer\_only),
all ten models.
Color peaks $\sim 4$--$8$ layers earlier than spatial in 9/10 models;
InternVL3-14B is the lone exception (same peak layer).}
\label{tab:color_peak}
\end{table}

We extend the framework to a non-spatial attribute via GQA color MCQs on
all ten models. CoT gates color transport in 10/10, including Qwen-7B
(open on spatial), whose color answer-only peak ($+0.082$ at L4)
collapses to $0.000$ under CoT.
Color peaks lie $\sim 4$--$8$ layers earlier than spatial peaks
(Table~\ref{tab:color_peak}), consistent with color being a shallower
visual attribute.

\paragraph{Amplitude as a boundary condition.}
Why does Qwen-7B flip between open (spatial) and fully suppressed (color)?
Across the ten models, gating strength is not monotone in encoding amplitude:
Janus encodes at $+0.325$ yet is fully suppressed, while Qwen-7B at $+0.156$ is not,
so there is no universal amplitude threshold across architectures.
Within Qwen-7B alone, however, lowering the encoding amplitude
flips the outcome: on spatial tasks (GQA $+0.137$; RefCOCO $+0.156$),
CoT does not gate; on color ($+0.082$), gating is complete.
The model, pipeline, and intervention are identical;
only the encoding amplitude differs.
This is consistent with a model-specific threshold:
amplitude can flip a single architecture's gating behaviour
even when no universal threshold separates architectures.

\section{Related Work}
\label{sec:related}

\paragraph{VLM spatial probing and grounding failures.}
VLMs are known to encode rich visual and spatial properties internally
while failing to use them at output
\citep{nooralahzadeh2026arbitration, liu2026seeingnotbelieving, asadi2026mirage, li2025groundingme, kamath2023whatsup, liu2023vsr}.
Closest to our setup, \citet{kang2026linearmechanisms} isolate linear
spatial-ID directions, with concurrent probes on fine-tuning origins
\citep{naghashyar2026spatialfeatures}
and the encoder/backbone split \citep{cui2026dualmechanisms}.
Broader probes cover grounding, hidden representations, corruption, and
storage--transfer
\citep{yu2025howmllms, liu2025visualrepresentations, golovanevsky2025vlmsnotice, basu2024informationstorage};
\citet{wu2026howvisionbecomeslanguage} decompose visual-to-language flow
via PID and attention knockouts.
These are largely static probes of what is encoded;
we take encoding as given and ask about transport.

\paragraph{Causal diagnostics and intervention.}
Selective heads and grounding circuits have been mapped via attention
inspection \citep{ma2026attentioninspace, chen2025whyspatialhard,
bi2025visualperception, kang2025fewheads, basile2025headpursuit},
but attention overlap is correlational.
Activation patching, causal mediation, and representation engineering
trace or steer LM computation via direction shifts
\citep{meng2022locating, geva2023dissecting, wang2023ioi, vig2020causalmediation, zou2023repeng, turner2023activation, li2023inference},
and causal abstraction \citep{geiger2024das} formalizes such groupings.
Our direction patching shares this primitive but localizes it across
layer, position, and prompt to map transport rather than steer behavior.

\paragraph{CoT and visual grounding prompts.}
CoT \citep{wei2022cot, kojima2022zeroshotcot} is the default for complex
reasoning, but \citet{kancheti2026cotdegrades} report that text CoT
\emph{degrades} VLM spatial accuracy and
\citet{mehrafarin2026cotfailshiddenstates} find correct solutions still
recoverable from hidden states; neither localizes where the pathway is blocked.
Visually grounded prompts
\citep{jiang2025vlmr3, zhang2025cofft, qin2025chain}
are typically framed as better recipes.
Recent decoding-time work also uses the temporal evolution of LVLM output
logits to preserve visual grounding: Residual Decoding aggregates logits
from semantically stable historical steps to provide history-aware guidance
against language-prior hallucinations \citep{chen2026resdec}.
Our analysis complements this line by localizing the upstream transport of
spatial evidence across residual-stream layers and token positions under
different prompt interfaces.
We provide complementary transport evidence: text CoT suppresses
immediate obj\_word argmax-level transport while visual grounding keeps
that route open, and a prefix-token route remains available in many
suppressed models.

\enlargethispage{2\baselineskip}

\section{Conclusion}
\label{sec:conclusion}

This work traces when encoded spatial information becomes causally available for VLM answers.
Direction patching shows that spatial transport is conditional on layer depth, prompt type, and token position:
spatial-ID directions can be available before they produce answer-logit
effects,
text CoT often suppresses immediate object-word argmax-level transport,
and visually grounded prompts keep this pathway open.
Across models, these local effects form descriptive transport groupings,
with behavioral steering and a proof-of-concept head knockout providing
additional, architecture-dependent evidence.
Together, the results refine the encoding-grounding dissociation:
VLMs may encode spatial information, but whether they use it depends on
how the prompt changes the layer, position, and route through which
that information reaches the answer logits. The answer-step and
stepwise controls further show that an immediate argmax-level suppression
does not by itself establish trajectory-wide erasure.

\section*{Limitations}

Direction patching probes the linear subspace identified by
\citet{kang2026linearmechanisms}: it manipulates a class-conditioned
direction in the residual stream and reads the resulting change in
answer logits.
All claims of ``transport'', ``gating'', and ``closed pathway'' in this
paper therefore refer to causal influence \emph{along this linear subspace}
under the tested prompt state, layer, and token position. In particular,
``gated'' denotes suppression of the four-way answer-option argmax effect
below the descriptive threshold; it does not denote complete erasure of
the target-logit signal or of every later decoding route.
The crossfit protocol and noise baselines control for spurious effects
within the subspace, but direction patching is silent on transport that
may occur through non-linear pathways or through alternative directions
outside this subspace.
The Janus case (\S\ref{sec:variation:patterns}: $+0.403$ visual\_cot
$\Delta$argmax but $0$pp generate-mode accuracy change) is a concrete
example of this gap: the linear subspace is highly responsive to
patching, yet the architecture's actual generation appears to rely on a
distinct pathway that our method cannot reach.
A fuller account of gating in VLMs will likely require complementary
non-linear probes (e.g.\ sparse autoencoders, dictionary learning) in
addition to direction patching.

We test two visual attributes (spatial position and color)
in a four-way forced-choice format.
Free-form spatial descriptions, fine-grained localization,
and other attribute types (size, shape, material) remain untested.

All ten models are open-weight, ranging from 4B to 32B parameters.
Closed-source models and architectures above 32B may exhibit different transport patterns.

Generate-mode behavioral validation uses 128-token generation and a
single object-word intervention at the specified layer. The resulting
parsed-answer changes can reflect effects that propagate through later
autoregressive states, so they should not be read as a direct estimate of
the fixed-state next-token $\Delta$argmax.
Longer generation or multi-turn interaction could produce different accuracy patterns
that our single-pass evaluation does not capture.

The visual grounding prompts we test are manually designed.
Systematic optimization of prompt wording for maximal bypass
could yield stronger effects but is beyond the scope of this study.

Our head-level localization is proof-of-concept, not a complete circuit.
The InternVL3-8B knockout identifies a single CoT-specific gating head
(L27.H21) whose removal reliably reopens transport,
and a parallel LLaVA-OV scan rules out a comparable single head there.
Tracing the upstream features that activate H21,
the downstream computation that completes the gate,
and the architectural reasons different families concentrate versus
distribute the same function are open questions.

\bibliography{bib/anthology_subset,bib/custom}

\renewcommand{\topfraction}{0.92}
\renewcommand{\bottomfraction}{0.30}
\renewcommand{\textfraction}{0.05}
\renewcommand{\floatpagefraction}{0.60}
\setcounter{topnumber}{3}
\setcounter{bottomnumber}{1}
\setcounter{totalnumber}{4}
\setcounter{dbltopnumber}{3}
\renewcommand{\dbltopfraction}{0.92}
\renewcommand{\dblfloatpagefraction}{0.60}

\appendix

\section{Prompt Templates}
\label{sec:appendix:prompts}

All experiments in the paper hold the question stem fixed and vary only the
instruction suffix appended to it. This isolates the prompt as the
manipulated variable so that any change in transport (e.g.\ CoT gating in
\S\ref{sec:transport:cot}, visual bypass in \S\ref{sec:transport:visual})
can be attributed to the suffix rather than to differences in the upstream
visual or linguistic input.

Table~\ref{tab:prompt_templates} lists the verbatim suffix for each of the
seven prompt formats. The first five (\texttt{cot}, \texttt{brief\_reason},
\texttt{answer\_first}, \texttt{final\_tag}, \texttt{answer\_only}) span a
reasoning-length spectrum from full text CoT down to direct letter output,
and are used in the prompt ladder of \S\ref{sec:variation:patterns}.
The last two (\texttt{visual\_cot}, \texttt{visual\_direct}) are visual-
grounding prompts that introduce intermediate tokens but redirect the
reasoning chain to visual evidence, isolating reasoning \emph{format} from
token \emph{count}.

\begin{table*}[!t]
\centering
\footnotesize
\setlength{\tabcolsep}{8pt}
\renewcommand{\arraystretch}{1.05}
\begin{tabular}{lp{12cm}}
\toprule
Format & Instruction suffix \\
\midrule
\rowcolor{gray!8} \texttt{cot}           & ``Let's think step by step.'' \\
\rowcolor{gray!8} \texttt{brief\_reason} & ``Give a brief reason, then answer with the letter.'' \\
\rowcolor{gray!8} \texttt{answer\_first} & ``First, answer with only the letter (A--D). Then explain briefly.'' \\
\rowcolor{gray!8} \texttt{final\_tag}    & ``Think about it, then give your final answer as Final Answer: X'' \\
\rowcolor{gray!8} \texttt{answer\_only}  & ``Answer with only the letter (A--D).'' \\
\rowcolor{gray!8} \texttt{visual\_cot}   & ``First, carefully look at the image and describe where you see the object mentioned above. Then answer with only the letter.'' \\
\rowcolor{gray!8} \texttt{visual\_direct}& ``Look at the object in the image. Note its position relative to the center. Answer with only the letter (A--D).'' \\
\bottomrule
\end{tabular}
\caption{Verbatim instruction suffixes appended to a shared question stem.
The first five suffixes (\texttt{cot} through \texttt{answer\_only}) span
a reasoning-length spectrum and are used in the prompt ladder of
\S\ref{sec:variation:patterns}; the last two (\texttt{visual\_cot},
\texttt{visual\_direct}) are visual-grounding prompts evaluated in
\S\ref{sec:transport:visual}.}
\label{tab:prompt_templates}
\end{table*}

\section{Per-Model Peak $\Delta$argmax Values}
\label{sec:appendix:peak_values}

Figure~\ref{fig:cot_gating} summarises CoT gating as donut cards;
this section provides the underlying numerical values.
Table~\ref{tab:cot_gating} reports per-model peak $\Delta$argmax under CoT
and answer-only at the object-word position, together with the layer at
which each peak is attained.

Eight of ten models fall below the descriptive $0.01$ threshold: six have
CoT peak $\Delta$argmax exactly $0.000$, while LLaVA ($+0.006$) and
Qwen-32B ($+0.003$) are near-zero/borderline cases. These values support
the descriptive transport patterns in \S\ref{sec:variation:patterns},
not a validated classification system.
InternVL2.5 is \emph{partially} gated: CoT collapses from $+0.168$ to
$+0.042$, a $75\%$ reduction, but the residual is non-zero.
Qwen-7B is the only \emph{open} model: its CoT peak ($+0.120$)
remains close to the answer-only peak ($+0.156$).

The peak layer is itself informative. Most 7B--8B models peak in the
L8--L16 band, but Qwen-32B and Gemma-3-27B shift to L24--L32; this
late-layer shift in larger models also appears in
\S\ref{sec:transport:late} and Appendix~\ref{sec:appendix:qwen32b}.
The peak values in this table are the canonical numbers used in the
abstract, the descriptive grouping, and the threshold analysis of
\S\ref{sec:variation:boundaries}.

\begin{table}[!tbp]
\centering
\small
\setlength{\tabcolsep}{3pt}
\renewcommand{\arraystretch}{1.05}
\begin{tabular}{l r r c}
\toprule
Model & CoT & Ans-only & Gated? \\
\midrule
\rowcolor{tintQwen}   \mQwen{Qwen-7B}         & +0.120\,{\scriptsize L16} & +0.156\,{\scriptsize L16} & \statOpen{No}     \\
\rowcolor{tintIntern} \mIntern{InternVL2.5}    & +0.042\,{\scriptsize L12} & +0.168\,{\scriptsize L12} & \statPartial{Weak} \\
\midrule
\rowcolor{tintLLaVA}  \mLLaVA{LLaVA}           & +0.006\,{\scriptsize L10} & +0.168\,{\scriptsize L16} & \statGated{Yes} \\
\rowcolor{tintIntern} \mIntern{InternVL3-8B}   & 0.000\,{\scriptsize all}  & +0.236\,{\scriptsize L16} & \statGated{Yes} \\
\rowcolor{tintIntern} \mIntern{InternVL3-14B}  & 0.000\,{\scriptsize all}  & +0.097\,{\scriptsize L24} & \statGated{Yes} \\
\rowcolor{tintGemma}  \mGemma{Gemma-3-4B}      & 0.000\,{\scriptsize all}  & +0.178\,{\scriptsize L12} & \statGated{Yes} \\
\rowcolor{tintGemma}  \mGemma{Gemma-3-12B}     & 0.000\,{\scriptsize all}  & +0.097\,{\scriptsize L20} & \statGated{Yes} \\
\rowcolor{tintGemma}  \mGemma{Gemma-3-27B}     & 0.000\,{\scriptsize all}  & +0.081\,{\scriptsize L24} & \statGated{Yes} \\
\rowcolor{tintJanus}  \mJanus{Janus}           & 0.000\,{\scriptsize all}  & +0.325\,{\scriptsize L12} & \statGated{Yes} \\
\rowcolor{tintQwen}   \mQwen{Qwen-32B}        & +0.003\,{\scriptsize L32} & +0.089\,{\scriptsize L32} & \statGated{Yes} \\
\bottomrule
\end{tabular}
\caption{Peak obj\_word $\Delta$argmax under CoT and answer-only
(RefCOCO, $\alpha{=}5$), with the layer at which each peak is attained.
The sample count is $n{=}50$/quadrant for 7B/8B models and
$n{=}30$/quadrant for Qwen-32B, Gemma-3, and Janus; each source sample is
paired with the three non-source target quadrants. Top two rows (above the
midrule) are open or partially suppressed; the bottom eight rows fall at
or below $0.006$.}
\label{tab:cot_gating}
\end{table}

\section{Controls}
\label{sec:appendix:controls}

A positive $\Delta$argmax at the object-word position could in principle
arise from two trivial sources rather than from direction-specific
spatial transport: any sufficiently large perturbation of the residual
stream might bias the answer logits, or the model might be sensitive to
any change at the object-word token regardless of its semantic content.
We address both possibilities here.

\begin{figure}[!tbp]
\centering
\includegraphics[width=\columnwidth]{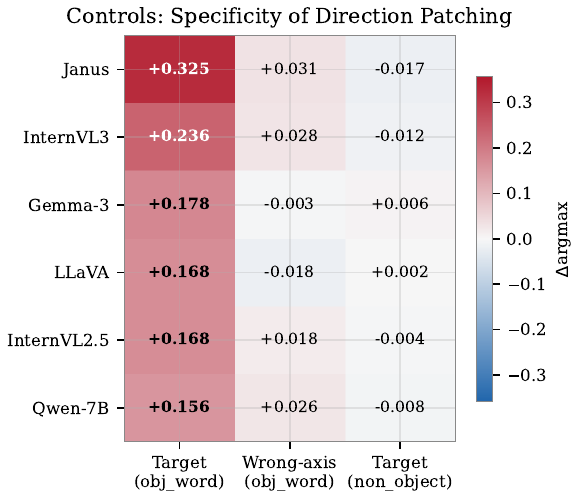}
\caption{Wrong-axis and non-object patching produce near-zero $\Delta$argmax,
confirming direction and position specificity.}
\label{fig:app_controls}
\end{figure}

Figure~\ref{fig:app_controls} and Table~\ref{tab:controls_full} report two
control conditions matched in $L_2$ norm to the target patch.
\emph{Wrong-axis} patches inject the perpendicular spatial direction at the
correct token, holding norm constant. \emph{Non-object} patches inject the
correct direction at a non-object token (a content word that is not the
referent). Both controls produce $\Delta$argmax within $\pm 0.03$ across
all ten models, an order of magnitude below the matched target direction
(e.g.\ Janus target $+0.325$ vs.\ wrong-axis $+0.031$ vs.\ non-obj $-0.017$).

Together, these controls rule out the two trivial explanations: the
target effect requires both the correct spatial direction and the correct
object-word position. The same axis-specific behaviour also holds on
COCO-Spatial (Appendix~\ref{sec:appendix:coco_spatial}), where horizontal
centroids do not shift vertical answers and vice versa.

\begin{table}[!tbp]
\centering
\footnotesize
\setlength{\tabcolsep}{2.5pt}
\begin{tabular}{lrrrr}
\toprule
Model & Lyr & Target & Wrong & Non-obj \\
\midrule
\rowcolor{tintQwen} \mQwen{Qwen-7B}       & 16 & +.156 & +.026 & $-.008$ \\
\rowcolor{tintLLaVA} \mLLaVA{LLaVA}         & 16 & +.168 & $-.018$ & +.002 \\
\rowcolor{tintIntern} \mIntern{IV2.5}         & 12 & +.168 & +.018 & $-.004$ \\
\rowcolor{tintIntern} \mIntern{IV3-8B}        & 16 & +.236 & +.028 & $-.012$ \\
\rowcolor{tintIntern} \mIntern{IV3-14B}       & 24 & +.097 & +.006 & +.003 \\
\rowcolor{tintGemma} \mGemma{Gem-4B}        & 12 & +.178 & $-.003$ & +.006 \\
\rowcolor{tintGemma} \mGemma{Gem-12B}       & 20 & +.097 & $-.006$ & $-.017$ \\
\rowcolor{tintGemma} \mGemma{Gem-27B}       & 24 & +.081 & $-.011$ & +.000 \\
\rowcolor{tintJanus} \mJanus{Janus}         & 12 & +.325 & +.031 & $-.017$ \\
\bottomrule
\end{tabular}
\caption{Direction and position controls (RefCOCO, ans-only, $\alpha{=}5$).
Wrong: perpendicular direction. Non-obj: correct direction at a non-object token.}
\label{tab:controls_full}
\end{table}

\section{Bootstrap Confidence Intervals}
\label{sec:appendix:ci}

The main text reports point estimates for $\Delta$argmax and target logit
gain. We additionally report sample-cluster bootstrap CIs
(2000 resamples; resampling unit is the sample id rather than the
sample--target pair) for the original four models at their canonical
layers (RefCOCO, $\alpha{=}5$, $n{=}50$/quadrant). CIs use heldout
centroids within each bootstrap fold, so no in-fold leakage inflates
significance.

Table~\ref{tab:ci_objword} reports CIs for obj\_word $\Delta$argmax.
For ans-only the CIs are tight and exclude zero by a wide margin
(e.g.\ Qwen-7B $[+.122,+.192]$). For CoT, LLaVA's lower bound touches
zero ($[+.000,+.006]$), and IV3-8B's CI is exactly
$[+.000,+.000]$. These intervals support the distinction between exact
zero and borderline near-zero cases; they do not turn the descriptive
threshold into a universal significance test.

Table~\ref{tab:ci_logit_gain} reports CIs for target logit gain (the raw
shift in the target-letter logit, before argmax). The relevant contrast
is between argmax and logit gain. Even when $\Delta$argmax is exactly zero (IV3-8B
CoT), the logit gain is positive ($+0.421$, $[+.341,+.501]$). The
direction is still pushing the target logit upward; it is simply not
enough to flip the argmax. The tested CoT interface therefore shows a
\emph{threshold} phenomenon (directional influence on the logit
distribution survives while argmax-level transport is suppressed), not
complete signal erasure. This distinction matters for the head-knockout analysis
(\S\ref{sec:variation:validation}): recovering even small amounts of
$\Delta$argmax is meaningful because the underlying logit signal is not
zeroed out.

\begin{table}[!tbp]
\centering
\footnotesize
\setlength{\tabcolsep}{2pt}
\begin{tabular}{llrrl}
\toprule
Model & Pmt & Lyr & $\Delta_{\mathrm{argmax}}$ & 95\% CI \\
\midrule
\rowcolor{tintQwen} \mQwen{Qwen-7B}  & ans & 16 & +.156 & $[+.122,+.192]$ \\
\rowcolor{tintQwen} \mQwen{Qwen-7B}  & cot & 16 & +.120 & $[+.090,+.150]$ \\
\rowcolor{tintLLaVA} \mLLaVA{LLaVA}    & ans & 16 & +.168 & $[+.134,+.202]$ \\
\rowcolor{tintLLaVA} \mLLaVA{LLaVA}    & cot & 16 & +.002 & $[+.000,+.006]$ \\
\rowcolor{tintIntern} \mIntern{IV2.5}    & ans & 12 & +.168 & $[+.134,+.204]$ \\
\rowcolor{tintIntern} \mIntern{IV2.5}    & cot & 12 & +.042 & $[+.026,+.060]$ \\
\rowcolor{tintIntern} \mIntern{IV3-8B}   & ans & 16 & +.236 & $[+.192,+.279]$ \\
\rowcolor{tintIntern} \mIntern{IV3-8B}   & cot & 16 & +.000 & $[+.000,+.000]$ \\
\bottomrule
\end{tabular}
\caption{95\% sample-cluster bootstrap CIs (2000 resamples,
$n{=}50$/quadrant) for obj\_word $\Delta$argmax under each prompt at
each model's canonical layer (RefCOCO, $\alpha{=}5$).
IV3-8B CoT is exactly zero with zero-width CI; LLaVA CoT lower bound
touches zero.}
\label{tab:ci_objword}
\end{table}

\begin{table}[!tbp]
\centering
\footnotesize
\setlength{\tabcolsep}{2pt}
\begin{tabular}{llrrrl}
\toprule
Model & Pmt & Lyr & gain & 95\% CI \\
\midrule
\rowcolor{tintQwen} \mQwen{Qwen-7B}  & ans & 16 & +1.252 & $[+1.130,+1.386]$ \\
\rowcolor{tintQwen} \mQwen{Qwen-7B}  & cot & 16 & +0.500 & $[+0.417,+0.589]$ \\
\rowcolor{tintLLaVA} \mLLaVA{LLaVA}    & ans & 16 & +0.855 & $[+0.753,+0.964]$ \\
\rowcolor{tintLLaVA} \mLLaVA{LLaVA}    & cot & 16 & +0.146 & $[+0.075,+0.217]$ \\
\rowcolor{tintIntern} \mIntern{IV2.5}    & ans & 12 & +1.527 & $[+1.367,+1.679]$ \\
\rowcolor{tintIntern} \mIntern{IV2.5}    & cot & 12 & +0.311 & $[+0.263,+0.362]$ \\
\rowcolor{tintIntern} \mIntern{IV3-8B}   & ans & 16 & +1.850 & $[+1.676,+2.028]$ \\
\rowcolor{tintIntern} \mIntern{IV3-8B}   & cot & 16 & +0.421 & $[+0.341,+0.501]$ \\
\bottomrule
\end{tabular}
\caption{95\% bootstrap CIs for the raw shift in the target-letter logit
($n{=}50$/quadrant, 2000 resamples, RefCOCO, $\alpha{=}5$).
Target logit gain is positive even when $\Delta$argmax is zero
(IV3-8B CoT $+0.421$), showing that CoT gating suppresses the argmax
flip rather than erasing the directional signal.}
\label{tab:ci_logit_gain}
\end{table}

\section{Alpha Sweep}
\label{sec:appendix:alpha}

If CoT gating were merely a small effect size, scaling the patch
strength $\alpha$ upward should eventually push the CoT $\Delta$argmax
above zero. We test this with an $\alpha$ sweep at both intervened
positions and compare the dose--response curve of answer-only to that
of CoT for the same model and layer.

\begin{figure*}[!t]
\centering
\includegraphics[width=0.95\textwidth]{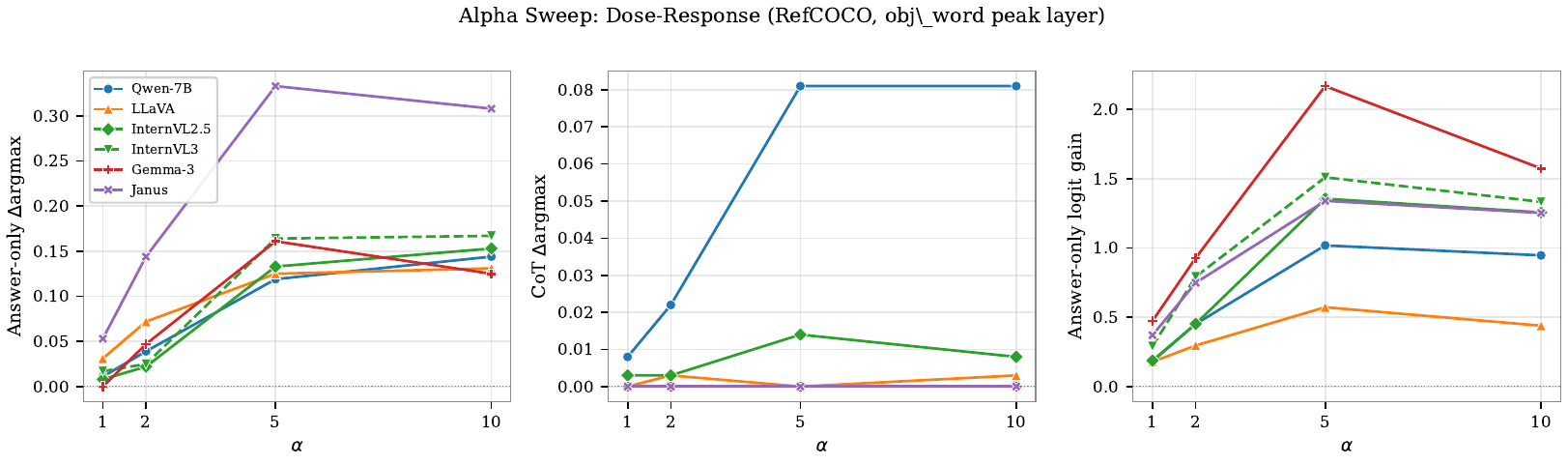}
\caption{Alpha sweep at obj\_word (RefCOCO, ten models).
Answer-only: increasing dose-response across the tested range. CoT:
near-zero for the suppressed majority, with LLaVA and Qwen-32B as
borderline cases.}
\label{fig:app_alpha}
\end{figure*}

We sweep $\alpha \in \{1,2,5,10\}$ at obj\_word
(Table~\ref{tab:alpha_objword}) and at prefix\_last
(Table~\ref{tab:alpha_prefix}). The qualitative pattern is the same at
both positions and is summarised in Figure~\ref{fig:app_alpha} for
obj\_word: answer-only is monotonic and grows with $\alpha$
(e.g.\ Qwen-7B $+.011 \!\to\! +.144$, IV3-8B $+.017 \!\to\! +.167$),
while the suppressed models remain at or near zero across the sweep.
LLaVA and Qwen-32B stay in the near-zero range, while the larger
non/partially suppressed effects occur for Qwen-7B and InternVL2.5.
The sweep therefore supports the same descriptive grouping without
turning the $0.01$ threshold into a claim of exact equality.

The prefix\_last sweep adds two specific observations. Qwen-7B CoT
$\Delta$argmax is non-monotonic in $\alpha$, peaking at $\alpha{=}5$
($+.356$) and collapsing at $\alpha{=}10$ ($+.069$), suggesting that the
prefix\_last pathway saturates earlier than the obj\_word pathway.
InternVL3-14B is the only model with an exact-zero pattern at prefix\_last
under both prompts at every $\alpha$; this is consistent with its
non-relocating transport pattern in
\S\ref{sec:variation:patterns}. The $\alpha{=}5$ value used throughout
the main text sits comfortably in the linear regime for both positions,
justifying our choice of a single canonical $\alpha$.

\begin{table*}[!t]
\centering
\footnotesize
\setlength{\tabcolsep}{6pt}
\begin{tabular}{lrrrrr}
\toprule
Model & Lyr & $\alpha{=}1$ & $\alpha{=}2$ & $\alpha{=}5$ & $\alpha{=}10$ \\
\midrule
\multicolumn{6}{l}{\textit{Answer-only}} \\
\rowcolor{tintQwen} \mQwen{Qwen-7B}   & 16 & +.011 & +.039 & +.119 & +.144 \\
\rowcolor{tintLLaVA} \mLLaVA{LLaVA}     & 16 & +.031 & +.072 & +.125 & +.131 \\
\rowcolor{tintIntern} \mIntern{IV2.5}     & 12 & +.008 & +.022 & +.133 & +.153 \\
\rowcolor{tintIntern} \mIntern{IV3-8B}    & 16 & +.017 & +.025 & +.164 & +.167 \\
\rowcolor{tintGemma} \mGemma{Gem-4B}    & 12 & .000  & +.047 & +.161 & +.125 \\
\rowcolor{tintJanus} \mJanus{Janus}     & 12 & +.053 & +.144 & +.333 & +.308 \\
\rowcolor{tintQwen} \mQwen{Qwen-32B}  & 32 & +.008 & +.008 & +.083 & +.294 \\
\rowcolor{tintIntern} \mIntern{IV3-14B}   & 16 & $-.006$ & .000 & +.014 & +.075 \\
\rowcolor{tintGemma} \mGemma{Gem-12B}   & 16 & +.003 & +.003 & +.036 & +.019 \\
\rowcolor{tintGemma} \mGemma{Gem-27B}   & 24 & +.022 & +.028 & +.089 & +.189 \\
\midrule
\multicolumn{6}{l}{\textit{CoT}} \\
\rowcolor{tintQwen} \mQwen{Qwen-7B}   & 16 & +.008 & +.022 & +.081 & +.081 \\
\rowcolor{tintLLaVA} \mLLaVA{LLaVA}     & 16 & .000  & +.003 & .000  & +.003 \\
\rowcolor{tintIntern} \mIntern{IV2.5}     & 12 & +.003 & +.003 & +.014 & +.008 \\
\rowcolor{tintIntern} \mIntern{IV3-8B}    & 16 & .000  & .000  & .000  & .000  \\
\rowcolor{tintGemma} \mGemma{Gem-4B}    & 12 & .000  & .000  & .000  & .000  \\
\rowcolor{tintJanus} \mJanus{Janus}     & 12 & .000  & .000  & .000  & .000  \\
\rowcolor{tintQwen} \mQwen{Qwen-32B}  & 32 & .000  & .000  & +.006 & .000 \\
\rowcolor{tintIntern} \mIntern{IV3-14B}   & 16 & .000  & .000  & .000  & .000 \\
\rowcolor{tintGemma} \mGemma{Gem-12B}   & 16 & .000  & .000  & .000  & .000 \\
\rowcolor{tintGemma} \mGemma{Gem-27B}   & 24 & .000  & .000  & .000  & .000 \\
\bottomrule
\end{tabular}
\caption{Obj\_word $\Delta$argmax across $\alpha \in \{1,2,5,10\}$ at each
model's canonical obj\_word layer (RefCOCO).
Answer-only grows across the dose range; the suppressed models remain at
or near zero. LLaVA and Qwen-32B are borderline near-zero cases, while
Qwen-7B and InternVL2.5 retain larger CoT effects.}
\label{tab:alpha_objword}
\end{table*}

\begin{table*}[!t]
\centering
\footnotesize
\setlength{\tabcolsep}{6pt}
\begin{tabular}{lrrrrr}
\toprule
Model & Lyr & $\alpha{=}1$ & $\alpha{=}2$ & $\alpha{=}5$ & $\alpha{=}10$ \\
\midrule
\multicolumn{6}{l}{\textit{Answer-only}} \\
\rowcolor{tintQwen} \mQwen{Qwen-7B}   & 20 & +.014 & +.047 & +.150 & +.275 \\
\rowcolor{tintLLaVA} \mLLaVA{LLaVA}     & 20 & +.014 & +.031 & +.075 & +.214 \\
\rowcolor{tintIntern} \mIntern{IV2.5}     & 24 & +.003 & +.008 & +.017 & +.039 \\
\rowcolor{tintIntern} \mIntern{IV3-8B}    & 20 & +.008 & +.025 & +.097 & +.461 \\
\rowcolor{tintIntern} \mIntern{IV3-14B}   & 36 & +.028 & +.044 & +.086 & +.344 \\
\rowcolor{tintGemma} \mGemma{Gem-4B}    & 24 & .000  & +.031 & +.050 & $-.025$ \\
\rowcolor{tintGemma} \mGemma{Gem-12B}   & 30 & +.008 & +.014 & +.033 & +.006 \\
\rowcolor{tintGemma} \mGemma{Gem-27B}   & 40 & +.011 & +.017 & +.056 & +.069 \\
\rowcolor{tintJanus} \mJanus{Janus}     & 20 & +.039 & +.094 & +.178 & +.228 \\
\rowcolor{tintQwen} \mQwen{Qwen-32B}  & 55 & .000  & .000  & +.008 & +.003 \\
\midrule
\multicolumn{6}{l}{\textit{CoT}} \\
\rowcolor{tintQwen} \mQwen{Qwen-7B}   & 20 & +.072 & +.211 & +.356 & +.069 \\
\rowcolor{tintLLaVA} \mLLaVA{LLaVA}     & 20 & .000  & .000  & +.008 & +.028 \\
\rowcolor{tintIntern} \mIntern{IV2.5}     & 24 & +.008 & +.006 & +.022 & +.100 \\
\rowcolor{tintIntern} \mIntern{IV3-8B}    & 20 & .000  & +.006 & +.033 & +.097 \\
\rowcolor{tintIntern} \mIntern{IV3-14B}   & 36 & .000  & .000  & .000  & .000 \\
\rowcolor{tintGemma} \mGemma{Gem-4B}    & 24 & .000  & .000  & +.006 & $-.003$ \\
\rowcolor{tintGemma} \mGemma{Gem-12B}   & 30 & .000  & .000  & $-.056$ & +.008 \\
\rowcolor{tintGemma} \mGemma{Gem-27B}   & 40 & .000  & .000  & $-.003$ & +.008 \\
\rowcolor{tintJanus} \mJanus{Janus}     & 20 & .000  & .000  & .000  & +.064 \\
\rowcolor{tintQwen} \mQwen{Qwen-32B}  & 55 & +.017 & +.019 & $-.014$ & $-.022$ \\
\bottomrule
\end{tabular}
\caption{Prefix\_last alpha sweep (RefCOCO).
Qwen-7B CoT non-monotonic (peak $\alpha{=}5$: $+.356$). IV3-14B: hard-zero at prefix\_last.}
\label{tab:alpha_prefix}
\end{table*}

\section{Prompt Ladder at Prefix\_last}
\label{sec:appendix:ladder}

\S\ref{sec:variation:patterns} distinguishes the \emph{prompt-selective}
pattern (Qwen-32B) from the suppressed--relocating pattern (LLaVA, IV2.5,
IV3-8B/14B) based on whether the prefix\_last pathway responds uniformly
across prompt formats. This appendix supplies the underlying data.
Holding the patch position fixed at prefix\_last and the layer at each
model's prefix\_last peak, we vary the prompt suffix across the
reasoning-length ladder of Appendix~\ref{sec:appendix:prompts}.

Figure~\ref{fig:app_ladder} and Table~\ref{tab:ladder} reveal three
prefix\_last behaviours.
The \emph{prompt-stable} pattern is exemplified by Qwen-7B at L28:
$\Delta$argmax stays near $0.80$ across all five formats, so the
prefix\_last route is open regardless of prompt.
The \emph{prompt-selective} pattern is exemplified by Qwen-32B at L55:
$\Delta$argmax reaches $+0.431$ under \texttt{final\_tag} but only
$+0.019$ under \texttt{answer\_only}, a $23\times$ swing that tracks
whether the prompt creates an explicit answer-anchor token.
The \emph{near-zero} pattern covers IV3-14B, Gem-12B and Gem-27B, which
remain at zero across every layer and every prompt at prefix\_last,
consistent with their non-relocating transport pattern.
LLaVA and IV2.5 sit in between: the prefix\_last route opens under
reasoning-style prompts (CoT, Brief, Ans-1st) but does not match
Qwen-7B's universal-open behaviour.
This prompt ladder is the data underlying the four descriptive transport
patterns
in \S\ref{sec:variation:patterns}.

\begin{figure*}[!t]
\centering
\includegraphics[width=0.95\textwidth]{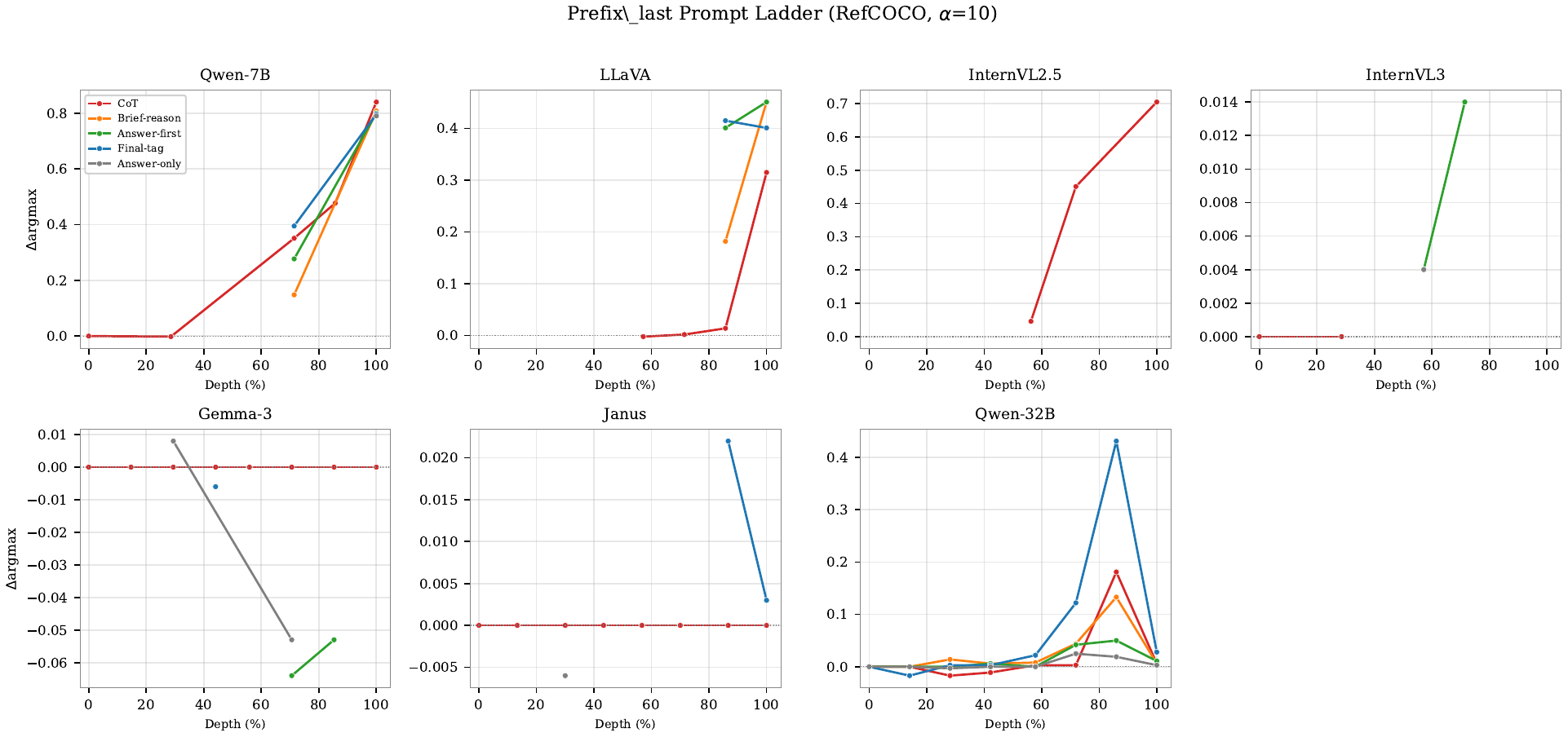}
\caption{Prompt ladder at prefix\_last (RefCOCO, $\alpha{=}10$).
Qwen-7B prompt-stable; Qwen-32B prompt-selective.}
\label{fig:app_ladder}
\end{figure*}

\begin{table*}[!t]
\centering
\footnotesize
\setlength{\tabcolsep}{8pt}
\begin{tabular}{llrrrrr}
\toprule
Model & Lyr & CoT & Brief & Ans-1st & Final & Ans \\
\midrule
\rowcolor{tintQwen}   \mQwen{Qwen-7B}  & 20 & +.351 & +.148 & +.277 & +.395 & --- \\
\rowcolor{tintQwen}   \mQwen{Qwen-7B}  & 28 & +.840 & +.808 & +.800 & +.796 & +.790 \\
\midrule
\rowcolor{tintQwen}   \mQwen{Qwen-32B} & 46 & +.003 & +.044 & +.042 & +.122 & +.025 \\
\rowcolor{tintQwen}   \mQwen{Qwen-32B} & 55 & +.181 & +.133 & +.050 & +.431 & +.019 \\
\midrule
\rowcolor{tintLLaVA}  \mLLaVA{LLaVA}   & 24 & +.014 & +.182 & +.401 & +.415 & --- \\
\rowcolor{tintLLaVA}  \mLLaVA{LLaVA}   & 28 & +.315 & +.451 & +.451 & +.401 & --- \\
\midrule
\rowcolor{tintIntern} \mIntern{IV2.5}  & 23 & +.451 & --- & --- & --- & --- \\
\rowcolor{tintIntern} \mIntern{IV2.5}  & 32 & +.705 & --- & --- & --- & --- \\
\midrule
\rowcolor{tintIntern} \mIntern{IV3-8B} & 16--20 & +.004 & +.004 & +.004 & --- & +.004 \\
\rowcolor{tintGemma}  \mGemma{Gem-4B}  & 0--34  & 0 & $-.064$ & $-.053$ & $-.006$ & 0 \\
\rowcolor{tintJanus}  \mJanus{Janus}   & 0--30  & 0 & +.003 & +.003 & +.022 & 0 \\
\midrule
\multicolumn{7}{l}{\textit{All zero across layers and prompts:} IV3-14B, Gem-12B, Gem-27B} \\
\bottomrule
\end{tabular}
\caption{Prefix\_last $\Delta$argmax under five prompt suffixes at each
model's prefix\_last layer (RefCOCO, $\alpha{=}10$).
Qwen-7B is prompt-stable; Qwen-32B is prompt-selective (final\_tag
$+0.431$ vs.\ ans-only $+0.019$); IV3-14B, Gem-12B, Gem-27B are zero
at every layer and every prompt. Dashes: condition not run.}
\label{tab:ladder}
\end{table*}

\section{Generate-Mode Accuracy}
\label{sec:appendix:accuracy}

The fixed-state transport results use next-token logit $\Delta$argmax at
$t{=}0$. To test whether an intervention at that state can influence the
actual decoded answer, we run $128$-token greedy generation under the same
direction patches and compare parsed-answer accuracy before and after the
intervention. These are related measurements with different denominators
and readout times.

Table~\ref{tab:accuracy} shows that steering toward incorrect quadrants
under answer-only drops accuracy by $8$ to $21$pp in nine of ten models
(Qwen-7B $74.2\% \!\to\! 53.6\%$; IV3-8B $77.5\% \!\to\! 56.9\%$;
Gem-12B $50.8\% \!\to\! 32.8\%$). Under text CoT the picture
splits: models whose CoT baselines have collapsed to near-chance
(Qwen-32B $9.2\%$, IV2.5 $4.2\%$) show $\leq 2.2$pp further drop,
whereas models with intact CoT baselines (LLaVA $55.8\%$, IV3-14B
$60.8\%$, Gem-12B $45.8\%$) still show $13$ to $16$pp drops despite
near-zero immediate obj\_word $\Delta$argmax. An obj\_word perturbation
can affect later autoregressive states and the final parsed answer even
when it does not change the first answer-option argmax; the independent
prefix\_last experiments provide the position-level route analysis.

Two cases deserve note. Visual\_cot steering produces sizable accuracy
drops for many models, while IV2.5, Gemma-3, and Janus show weaker or
absent changes. This is consistent with visual grounding preserving an
active behavioral pathway in many, but not all, generation tests.
Janus is the principal exception: its visual\_cot $\Delta$argmax is the
strongest in the study ($+0.403$), yet generate-mode accuracy moves by
$\le 3$pp under any prompt. The unified vision--language architecture
appears to route generation through a pathway decoupled from the linear
spatial subspace we patch; this is the sharpest encoding-without-behavior
boundary case in our data.

\begin{table*}[!t]
\centering
\footnotesize
\setlength{\tabcolsep}{4pt}
\begin{tabular}{llclrrrr}
\toprule
Model & Pmt & Lyr & Pos & Base & Steer & $\Delta$acc & $\Delta_{\mathrm{parse}}$ \\
\midrule
\rowcolor{tintQwen}   \mQwen{Qwen-7B}     & ans & 16 & obj & 74.2 & 53.6 & $-$20.6 & +.125 \\
\rowcolor{tintQwen}                       & vis & 16 & obj & 72.5 & 57.8 & $-$14.7 & +.114 \\
\rowcolor{tintQwen}                       & cot & 16 & obj & 20.0 & 18.6 & $-$1.4  & $-$.006 \\
\midrule
\rowcolor{tintQwen}   \mQwen{Qwen-32B}    & ans & 32 & obj & 77.5 & 61.1 & $-$16.4 & +.072 \\
\rowcolor{tintQwen}                       & vis & 32 & obj & 70.8 & 54.7 & $-$16.1 & +.044 \\
\rowcolor{tintQwen}                       & cot & 32 & obj & 9.2  & 6.9  & $-$2.2  & $-$.019 \\
\midrule
\rowcolor{tintLLaVA}  \mLLaVA{LLaVA}      & ans & 16 & obj & 55.8 & 36.9 & $-$18.9 & +.106 \\
\rowcolor{tintLLaVA}                      & vis & 16 & obj & 57.5 & 36.9 & $-$20.6 & +.131 \\
\rowcolor{tintLLaVA}                      & cot & 16 & obj & 55.8 & 40.6 & $-$15.3 & +.117 \\
\midrule
\rowcolor{tintIntern} \mIntern{IV2.5}     & ans & 16 & obj & 69.2 & 60.3 & $-$8.9  & +.053 \\
\rowcolor{tintIntern}                     & vis & 16 & obj & 67.5 & 60.6 & $-$6.9  & +.053 \\
\rowcolor{tintIntern}                     & cot & 16 & obj & 4.2  & 4.4  & +0.3    & +.019 \\
\midrule
\rowcolor{tintIntern} \mIntern{IV3-8B}    & ans & 16 & obj & 77.5 & 56.9 & $-$20.6 & +.164 \\
\rowcolor{tintIntern}                     & vis & 16 & obj & 78.3 & 57.2 & $-$21.1 & +.122 \\
\rowcolor{tintIntern}                     & cot & 16 & obj & 33.3 & 27.5 & $-$5.8  & +.064 \\
\midrule
\rowcolor{tintIntern} \mIntern{IV3-14B}   & ans & 20 & obj & 81.7 & 65.6 & $-$16.1 & +.072 \\
\rowcolor{tintIntern}                     & vis & 20 & obj & 82.5 & 64.4 & $-$18.1 & +.094 \\
\rowcolor{tintIntern}                     & cot & 20 & obj & 60.8 & 47.8 & $-$13.1 & +.067 \\
\midrule
\rowcolor{tintGemma}  \mGemma{Gem-4B}     & ans & 12 & obj & 40.0 & 26.7 & $-$13.3 & +.125 \\
\rowcolor{tintGemma}                      & vis & 12 & obj & 44.2 & 31.9 & $-$12.2 & +.078 \\
\rowcolor{tintGemma}                      & cot & 12 & obj & 30.0 & 26.7 & $-$3.3  & +.028 \\
\midrule
\rowcolor{tintGemma}  \mGemma{Gem-12B}    & ans & 18 & obj & 50.8 & 32.8 & $-$18.1 & +.139 \\
\rowcolor{tintGemma}                      & vis & 18 & obj & 50.8 & 38.9 & $-$11.9 & +.036 \\
\rowcolor{tintGemma}                      & cot & 18 & obj & 45.8 & 30.3 & $-$15.6 & +.061 \\
\midrule
\rowcolor{tintGemma}  \mGemma{Gem-27B}    & ans & 24 & obj & 55.8 & 46.4 & $-$9.4  & +.100 \\
\rowcolor{tintGemma}                      & vis & 24 & obj & 60.0 & 53.3 & $-$6.7  & +.067 \\
\rowcolor{tintGemma}                      & cot & 24 & obj & 55.8 & 49.4 & $-$6.4  & +.042 \\
\midrule
\rowcolor{tintJanus}  \mJanus{Janus}      & ans & 16 & obj & 50.0 & 50.0 & +0.0    & $-$.006 \\
\rowcolor{tintJanus}                      & vis & 16 & obj & 57.5 & 60.3 & +2.8    & $-$.014 \\
\rowcolor{tintJanus}                      & cot & 16 & obj & 41.7 & 41.1 & $-$0.6  & +.017 \\
\bottomrule
\end{tabular}
\caption{Full $128$-token greedy-generation accuracy (\%, RefCOCO,
$n{=}30$/quadrant, $120$ source samples and $360$ source--target
intervention records per cell) before steering (Base) and after steering
toward an incorrect quadrant (Steer). Pos: obj $=$ obj\_word.
$\Delta$acc is the change in parsed-answer accuracy in percentage points;
$\Delta_{\mathrm{parse}}$ is the intervention-minus-noise contrast in the
final parsed target rate. It is not the fixed-state next-token
$\Delta$argmax. Pmt: ans $=$ answer-only, vis $=$ visual\_cot, cot $=$
text CoT.}
\label{tab:accuracy}
\end{table*}

\section{GQA Spatial Layer Sweep}
\label{sec:appendix:gqa_spatial}

The RefCOCO-derived MCQs in the main text use referring expressions and
COCO images. GQA Spatial uses different images, scene graphs, and an
independent object distribution, so reproducing the transport map on GQA
tests whether the patterns are dataset-specific or properties of the
underlying models. Table~\ref{tab:gqa_spatial_full} reports the complete
layer-by-layer sweep for four representative models that span the four
descriptive transport patterns.

Three structural predictions from RefCOCO are preserved on GQA. First,
the suppression status is stable: LLaVA, IV3-8B and Janus remain
near-zero at obj\_word under CoT (peak $\le 0.003$), and Qwen-7B remains
open ($+.142$ under CoT). Second, peak layers are stable within
$\pm 8$ layers: IV3-8B's obj\_word peak is at L12 on GQA and L16 on
RefCOCO, well within the $8$-layer tolerance. Third, the relocation
pattern holds for the suppressed--relocating models: IV3-8B's prefix\_last
peaks at L20 with $+0.455$ (its largest cell), while obj\_word saturates
at L12 ($+0.112$). Janus stays non-relocating with negative prefix\_last
throughout.

The takeaway is that the transport map is a model property, not a quirk
of RefCOCO scene composition. The two minor deviations are consistent
with the boundary analysis in \S\ref{sec:variation:boundaries}: Gemma-3-27B
shows a weak prefix\_last effect on GQA ($+0.187$ at L40) not visible at
the RefCOCO ladder resolution, and Janus reproduces its sharp
non-relocating shape across both datasets, suggesting the architecture is
the controlling factor rather than the input distribution.

\begin{table*}[!t]
\centering
\footnotesize
\setlength{\tabcolsep}{6pt}
\begin{tabular}{llrrrrrrrr}
\toprule
Model & Cond & L0 & L4 & L8 & L12 & L16 & L20 & L24 & L28 \\
\midrule
\rowcolor{tintQwen}   \mQwen{Qwen-7B} & o,a & 0 & +.083 & \textbf{+.137} & +.095 & +.112 & +.002 & +.002 & 0 \\
\rowcolor{tintQwen}    & o,c & +.003 & +.083 & \textbf{+.142} & +.113 & +.100 & +.007 & $-.007$ & 0 \\
\rowcolor{tintQwen}    & p,a & $-.002$ & +.012 & +.055 & +.088 & +.267 & \textbf{+.322} & +.073 & $-.020$ \\
\rowcolor{tintQwen}    & p,c & +.005 & +.060 & +.163 & +.137 & \textbf{+.213} & +.185 & +.030 & $-.017$ \\
\midrule
\rowcolor{tintLLaVA}  \mLLaVA{LLaVA} & o,a & +.002 & +.085 & \textbf{+.117} & +.090 & +.050 & $-.003$ & 0 & 0 \\
\rowcolor{tintLLaVA}   & o,c & 0 & 0 & +.002 & \textbf{+.003} & +.002 & 0 & 0 & 0 \\
\rowcolor{tintLLaVA}   & p,a & 0 & +.025 & +.037 & +.085 & \textbf{+.347} & +.245 & +.055 & +.068 \\
\rowcolor{tintLLaVA}   & p,c & +.002 & +.040 & +.020 & +.042 & \textbf{+.182} & +.142 & +.010 & +.048 \\
\midrule
\rowcolor{tintIntern} \mIntern{IV3-8B} & o,a & +.003 & +.082 & +.082 & \textbf{+.112} & +.075 & +.012 & +.002 & 0 \\
\rowcolor{tintIntern}  & o,c & 0 & 0 & 0 & 0 & 0 & $-.002$ & 0 & 0 \\
\rowcolor{tintIntern}  & p,a & $-.002$ & $-.005$ & +.012 & +.177 & +.382 & \textbf{+.455} & +.180 & +.002 \\
\rowcolor{tintIntern}  & p,c & 0 & +.003 & +.008 & +.042 & \textbf{+.148} & +.098 & +.025 & +.018 \\
\midrule
\rowcolor{tintJanus}  \mJanus{Janus} & o,a & +.020 & +.085 & +.198 & \textbf{+.315} & +.002 & +.002 & 0 & +.002 \\
\rowcolor{tintJanus}   & o,c & 0 & 0 & 0 & 0 & 0 & 0 & 0 & 0 \\
\rowcolor{tintJanus}   & p,a & +.035 & $-.017$ & $-.075$ & $-.033$ & $-.130$ & $-.077$ & $-.098$ & $-.035$ \\
\rowcolor{tintJanus}   & p,c & 0 & +.010 & +.010 & +.012 & +.007 & $-.103$ & $-.057$ & 0 \\
\bottomrule
\end{tabular}
\caption{GQA Spatial layer sweep (4 models).
o $=$ obj\_word, p $=$ prefix\_last, a $=$ ans-only, c $=$ CoT. Bold: peak.
IV3-8B: relocation (obj L12, pfx L20) with position-specific gating.
Janus: obj peaks at L12; pfx negative.}
\label{tab:gqa_spatial_full}
\end{table*}

\section{GQA Color Sweep}
\label{sec:appendix:gqa_color}

We also test whether the transport apparatus applies to a non-spatial
visual attribute. GQA Color MCQs are constructed by the same procedure
as GQA Spatial, replacing the spatial question with a
``what color is the X'' question grounded in scene-graph color
annotations. The framework is held constant; only the attribute changes.

All ten models were evaluated on GQA Color, and CoT gates color transport
in 10/10 (compared with 8/10 on spatial; per-model peak layers in
Table~\ref{tab:color_peak}).
Table~\ref{tab:gqa_color} gives the full obj\_word layer sweep for the
six models with $\le 30$-layer LMs; peak layer summaries for the larger
variants (Qwen-32B, InternVL3-14B, Gemma-3-12B, Gemma-3-27B) are in
Table~\ref{tab:color_peak} of the main text.
Color peaks are systematically four to eight layers earlier than the
corresponding spatial peaks, consistent with color being a shallower
visual feature.
Peak amplitudes are also smaller on color (typical maxima
$+0.08$ to $+0.14$ for color vs.\ $+0.10$ to $+0.32$ for spatial),
reflecting a weaker color direction in our centroid construction.

These two effects together explain the Qwen-7B gating flip discussed in
\S\ref{sec:variation:boundaries}. Qwen-7B is open on spatial tasks,
but its color answer-only peak is only $+0.082$, below the threshold
needed to sustain CoT transport, and CoT color $\Delta$argmax collapses
to $0.000$. The architecture, pipeline, and intervention are identical;
only the encoding amplitude differs.

\begin{table*}[!t]
\centering
\footnotesize
\setlength{\tabcolsep}{10pt}
\begin{tabular}{lrrrrrrrr}
\toprule
Model & L0 & L4 & L8 & L12 & L16 & L20 & L24 & L28 \\
\midrule
\rowcolor{tintQwen} \mQwen{Qwen-7B} & 0 & \textbf{+.082} & +.040 & +.028 & +.010 & 0 & $-.002$ & 0 \\
\rowcolor{tintLLaVA} \mLLaVA{LLaVA}   & +.002 & \textbf{+.103} & +.062 & +.018 & +.007 & $-.002$ & +.002 & 0 \\
\rowcolor{tintIntern} \mIntern{IV2.5}   & +.003 & \textbf{+.102} & +.098 & +.023 & +.018 & +.005 & 0 & $-.002$ \\
\rowcolor{tintIntern} \mIntern{IV3-8B}  & +.002 & \textbf{+.107} & +.073 & +.033 & $-.002$ & 0 & 0 & 0 \\
\rowcolor{tintGemma} \mGemma{Gem-4B}  & +.047 & +.017 & \textbf{+.078} & +.018 & +.005 & 0 & $-.002$ & +.002 \\
\rowcolor{tintJanus} \mJanus{Janus}   & +.015 & +.077 & \textbf{+.082} & +.052 & +.002 & 0 & 0 & 0 \\
\bottomrule
\end{tabular}
\caption{Layer-wise $\Delta$argmax at obj\_word on GQA Color
(answer-only, $\alpha{=}5$, $n{=}50$/quadrant).
Bold: per-model peak layer. Color peaks at L4--L8, approximately four
layers earlier than the spatial-attribute peaks at L8--L16.}
\label{tab:gqa_color}
\end{table*}

\section{Visual Prompt Comparison}
\label{sec:appendix:visual}

\S\ref{sec:transport:visual} contrasts text CoT (\texttt{cot}) with visual
grounding (\texttt{visual\_cot}) and reports that the latter bypasses
obj\_word gating. We compare both visual prompts (\texttt{visual\_cot}
and the briefer \texttt{visual\_direct}) against text CoT and
answer\_only at both intervened positions, testing whether the bypass is
a property of visual grounding in general or a property of the specific
prompt wording.

Table~\ref{tab:visual_full} shows that both visual prompts bypass CoT
gating at obj\_word in every gated model: Vis-C and Vis-D both produce
positive $\Delta$argmax even when standard CoT is exactly zero
(e.g.\ IV3-8B Vis-C $+.164$, Vis-D $+.144$, CoT $.000$). The bypass is
not specific to one wording. A second observation is that visual prompts
sometimes open routes that even answer-only cannot:
Janus's Vis-C/Vis-D values ($+.403$/$+.397$) exceed its answer-only peak
($+.069$) by a factor of $\sim 6$, suggesting that visual grounding
prompts can amplify spatial transport beyond the canonical answer-only
baseline in models that route through visual-thinking pathways.

At prefix\_last, the bypass is also visible: Gem-4B reaches $+.206$
under Vis-C despite a null prefix\_last under both CoT and answer-only,
indicating that visual grounding can reopen a position-level route that
standard prompts leave this route near zero. Qwen-32B shows a similar pattern,
with prefix\_last Vis-C $+.217$ and Vis-D $+.200$, consistent with its
prompt-selective grouping in which the prefix\_last route depends on
prompt format.

Across families, the two visual prompts generally agree in direction and
magnitude: the Pearson correlation between Vis-C and Vis-D at obj\_word
is $r > 0.99$ across the ten models, and the largest Vis-C / Vis-D
discrepancy is Gem-12B ($+.064$ vs.\ $+.103$). This confirms that
visual grounding bypass is a property of the prompt class, not a quirk
of one template. Janus remains the most amplified case: its Vis-C and
Vis-D values at obj\_word ($+.403$ / $+.397$) also extend to
prefix\_last ($+.244$ / $+.203$), making it the only non-relocating
model with strong positive prefix\_last under visual prompts.

As an additional axis-specific control,
horizontal-axis centroids do not shift vertical-axis answers, and vice
versa (8 axis-transfer experiment files on RefCOCO), so the spatial
transport documented here is direction-specific rather than a generic
sensitivity to any centroid difference.

\begin{table*}[!t]
\centering
\footnotesize
\setlength{\tabcolsep}{8pt}
\begin{tabular}{lrrrr@{\hskip 18pt}lrrrr}
\toprule
\multicolumn{5}{c}{\textit{Obj\_word peak}} & \multicolumn{5}{c}{\textit{Prefix\_last peak}} \\
\midrule
Model & CoT & Vis-C & Vis-D & Ans & Model & CoT & Vis-C & Vis-D & Ans \\
\midrule
\rowcolor{tintQwen} \mQwen{Qwen-7B}   & +.081 & +.136 & +.111 & +.169 & \mQwen{Qwen-7B}   & --- & +.122 & +.169 & --- \\
\rowcolor{tintQwen} \mQwen{Qwen-32B}  & +.006 & +.117 & +.097 & +.089 & \mQwen{Qwen-32B}  & --- & +.217 & +.200 & --- \\
\rowcolor{tintLLaVA} \mLLaVA{LLaVA}     & .000  & +.103 & +.117 & +.158 & \mLLaVA{LLaVA}     & --- & +.108 & +.097 & --- \\
\rowcolor{tintIntern} \mIntern{IV2.5}     & .000  & +.153 & +.128 & +.114 & \mIntern{IV2.5}     & --- & +.058 & +.047 & --- \\
\rowcolor{tintIntern} \mIntern{IV3-8B}    & .000  & +.164 & +.144 & +.236 & \mIntern{IV3-8B}    & --- & +.144 & +.156 & --- \\
\rowcolor{tintIntern} \mIntern{IV3-14B}   & .000  & +.117 & +.108 & +.097 & \mIntern{IV3-14B}   & --- & +.108 & +.097 & --- \\
\rowcolor{tintGemma} \mGemma{Gem-4B}    & .000  & +.092 & +.100 & +.178 & \mGemma{Gem-4B}    & --- & +.206 & +.142 & --- \\
\rowcolor{tintGemma} \mGemma{Gem-12B}   & .000  & +.064 & +.103 & +.097 & \mGemma{Gem-12B}   & --- & +.131 & +.022 & --- \\
\rowcolor{tintGemma} \mGemma{Gem-27B}   & .000  & +.022 & +.061 & +.081 & \mGemma{Gem-27B}   & --- & +.022 & +.072 & --- \\
\rowcolor{tintJanus} \mJanus{Janus}     & .000  & +.403 & +.397 & +.069 & \mJanus{Janus}     & --- & +.244 & +.203 & --- \\
\bottomrule
\end{tabular}
\caption{Peak $\Delta$argmax under four prompt types at obj\_word (left) and
prefix\_last (right), RefCOCO, $\alpha{=}5$.
Vis-C $=$ visual\_cot, Vis-D $=$ visual\_direct, Ans $=$ answer\_only.
Both visual prompts produce positive $\Delta$argmax in every gated
model. Dashes: condition not run.}
\label{tab:visual_full}
\end{table*}

\section{Stepwise CoT Generation Probe}
\label{sec:appendix:stepwise}

The obj\_word gating result in \S\ref{sec:transport:cot} measures
$\Delta$argmax only at the first generation step. If CoT merely delays
rather than blocks the spatial pathway, the patched direction should
surface at some later step in the reasoning trace and shift the parsed
final answer. We test this with a stepwise probe.

We record per-step A/B/C/D logits at every greedy decoding step for
Qwen-7B and LLaVA at L16, $\alpha{=}5$, and report both the best
single-step target-logit gain over the trajectory and the final answer hit
rate (Table~\ref{tab:stepwise}). Object-word patching does move later-step
logits in absolute terms (best gain $+1.32$ at Qwen-7B step $65$;
$+2.88$ at LLaVA step $109$). These are non-trivial logit shifts, but
they do not by themselves establish a stable answer readout at those
steps.

What does not change, however, is the parsed answer. Across $32$, $64$
and $128$-step traces, the final-answer hit rate moves by at most
$+0.10$ over the noise control. The intervention pushes the target
logit upward, but rarely enough to flip the argmax against the rest of
the sequence's accumulated evidence. The probe therefore does not support
a simple ``first token versus later token'' explanation: later logit shifts
can occur without systematic parsed-answer recovery. It also does not
prove that every generated step has the same transport status. The
empty-answer fractions in the rightmost column
(intervention$/$noise) are matched, ruling out the explanation that the
intervention simply destabilises generation into unparseable output.

\begin{table*}[!t]
\centering
\footnotesize
\setlength{\tabcolsep}{10pt}
\begin{tabular}{lrrrrrc}
\toprule
Model & steps & $n$/q & best step & best $\Delta$argmax & final $\Delta$ & empty \\
\midrule
\rowcolor{tintQwen} \mQwen{Qwen-7B} & 32  & 20 & 19  & +.008 & +.017 & .82/.80 \\
\rowcolor{tintQwen} \mQwen{Qwen-7B} & 64  & 10 & 13  & +.008 & $-$.025 & .63/.48 \\
\rowcolor{tintQwen} \mQwen{Qwen-7B} & 128 & 20 & 65  & +.031 & +.025 & .41/.35 \\
\rowcolor{tintLLaVA} \mLLaVA{LLaVA}   & 32  & 20 & 3   & +.013 & +.042 & .25/.37 \\
\rowcolor{tintLLaVA} \mLLaVA{LLaVA}   & 64  & 10 & 60  & +.030 & +.025 & .07/.10 \\
\rowcolor{tintLLaVA} \mLLaVA{LLaVA}   & 128 & 20 & 109 & +.200 & +.104 & .02/.01 \\
\bottomrule
\end{tabular}
\caption{Stepwise CoT probe.
``best step'': generation index with maximal target-logit gain;
``best $\Delta$argmax'': target-option argmax-rate contrast at that step.
``final $\Delta$'': change in parsed-answer hit rate (intervention $-$ noise).
``empty'': fraction of unparseable outputs (intervention/noise).}
\label{tab:stepwise}
\end{table*}

\section{Answer-Step Timing Control}
\label{sec:appendix:answer_step}

The stepwise probe above follows every generated token, but it does not
force the readout to occur at the model's eventual answer step. We therefore
run a targeted timing control on RefCOCO. For each clean text-CoT
generation, we locate the step at which the model produces its answer
letter, reconstruct the prefix immediately before that letter, and read
the A/B/C/D logits there. We then apply the same object-word direction
patch and same-norm random-direction control at that answer step. The
direction, crossfit procedure, $\alpha{=}5$, and $128$-token budget are
otherwise unchanged.

Table~\ref{tab:answer_step} reports the target-logit gain on all samples
for which an answer step was found. Each source sample is paired with the
three non-source target quadrants, giving $240$, $177$, and $216$ records
for LLaVA, Qwen, and InternVL3, respectively. The target-logit gain is
positive from L8 through L20 in all three models and fades at L24. The
same sign pattern holds on the clean-correct subset, although the
secondary $\Delta$argmax is smaller and model-dependent. This control
addresses the possibility that the first fixed readout is mistimed without
claiming that all decoding steps share one transport profile.

\begin{table*}[!t]
\centering
\footnotesize
\setlength{\tabcolsep}{8pt}
\begin{tabular}{lrrrrrr}
\toprule
Model & answer-found $n$ & L8 & L12 & L16 & L20 & L24 \\
\midrule
\rowcolor{tintLLaVA} \mLLaVA{LLaVA-OV-7B} & 80 & +.100 & +.280 & +.430 & +.080 & $-.019$ \\
\rowcolor{tintQwen}  \mQwen{Qwen2.5-VL-7B} & 59 & +.221 & +.204 & +.309 & +.111 & $-.000$ \\
\rowcolor{tintIntern} \mIntern{InternVL3-8B} & 72 & +.649 & +.763 & +.538 & +.176 & $-.081$ \\
\bottomrule
\end{tabular}
\caption{Target-logit gain at the model's actual generated answer step
under an obj\_word direction patch (RefCOCO, text CoT, $\alpha{=}5$,
five-fold crossfit). Values use all answer-found samples; the answer step
is the position immediately before the generated A/B/C/D letter. The
corresponding $\Delta$argmax at L16 is $+.125$, $+.011$, and $-.005$ for
LLaVA, Qwen, and InternVL3, respectively, so the continuous gain is the
primary timing-control readout.}
\label{tab:answer_step}
\end{table*}

\section{Blank-Image and Full-CoT Sanity}
\label{sec:appendix:blank}

Two further controls address residual confounds.
\emph{Blank-image}: if the patched centroids were dominated by
linguistic priors rather than visual content, patching at a blank
canvas should still reproduce the original effect.
\emph{Full-CoT generation}: if the gating result depended on the
specific phrasing of our \texttt{cot} suffix, a longer or differently
worded CoT chain might still expose transport at later steps.

Table~\ref{tab:blank} shows that the blank-image substitution collapses
the original-image effect ($+.156$ for Qwen-7B, $+.168$ for LLaVA) to a
residual $\pm 0.04$ at $n{=}20$/q. This residual is within the noise
floor of the bootstrap CIs in Appendix~\ref{sec:appendix:ci} for the
same models. The direction we patch is therefore image-grounded, not
image-independent. The full-CoT rows are intentionally underpowered
($n{=}10$/q) and serve only as a cross-check against the more
informative $128$-step probe in Table~\ref{tab:stepwise}; both point in
the same direction.

\section{COCO-Spatial Axis Transfer}
\label{sec:appendix:coco_spatial}

The four-quadrant RefCOCO MCQs combine horizontal (left/right) and
vertical (above/below) information into a single answer letter, so the
quadrant centroid we patch in principle mixes both axes. To check that
each axis is independently encoded and transported, we construct a
parallel COCO-Spatial benchmark from 2-object captions and convert it
to two separate binary MCQs (left/right; above/below).

Table~\ref{tab:coco_spatial} reports L16, $n{=}50$/class, $\alpha{=}5$,
with 5-fold heldout centroids per axis.
Horizontal transport reproduces cleanly:
Qwen-7B reaches $\Delta$argmax $+.220$ under CoT and $+.210$ under
answer-only, mirroring its open RefCOCO behaviour, and LLaVA
shows the same CoT-zero / ans-positive split as on RefCOCO ($.000$
vs.\ $+.170$).
Vertical transport is systematically weaker than horizontal: both
models drop by roughly an order of magnitude on the above/below task
($+.070$ for Qwen-7B; $.000$ / $-.010$ for LLaVA).

This horizontal--vertical asymmetry is consistent with the broader
literature on VLM spatial reasoning, where left/right has stronger
linguistic and visual scaffolding than above/below \citep{kamath2023whatsup, liu2023vsr}.
For our purposes the relevant result is that each axis is
independently steerable, so the quadrant centroid we patch in RefCOCO is
not an artefact of axis confounding.

\begin{table}[!tbp]
\centering
\footnotesize
\setlength{\tabcolsep}{3pt}
\begin{tabular}{llrrrr}
\toprule
Model & Ax & CoT$_{\Delta}$ & Ans$_{\Delta}$ & CoT$_{g}$ & Ans$_{g}$ \\
\midrule
\rowcolor{tintQwen} \mQwen{Qwen-7B} & H & +.220 & +.210 & +1.04 & +0.34 \\
\rowcolor{tintLLaVA} \mLLaVA{LLaVA}   & H & .000  & +.170 & +0.44 & +0.40 \\
\rowcolor{tintQwen} \mQwen{Qwen-7B} & V & +.070 & +.060 & +0.35 & +0.17 \\
\rowcolor{tintLLaVA} \mLLaVA{LLaVA}   & V & .000  & $-.010$ & +0.16 & +0.06 \\
\bottomrule
\end{tabular}
\caption{Axis-specific transport on COCO-Spatial binary MCQs
($n{=}50$/class, $\alpha{=}5$, L16, 5-fold heldout centroids).
Ax: H $=$ horizontal (left/right), V $=$ vertical (above/below).
$\Delta$ $=$ $\Delta$argmax; $g$ $=$ target-letter logit gain.
Horizontal axis reproduces the RefCOCO CoT-zero / ans-positive pattern;
vertical axis is systematically weaker.}
\label{tab:coco_spatial}
\end{table}

\begin{table}[!tbp]
\centering
\footnotesize
\setlength{\tabcolsep}{2pt}
\begin{tabular}{llllrrr}
\toprule
Model & Pmt & Eval & Img & $n$/q & Contrast & gain \\
\midrule
\rowcolor{tintQwen} \mQwen{Qwen-7B} & cot & gen  & orig  & 10 & $-.033$ & --- \\
\rowcolor{tintLLaVA} \mLLaVA{LLaVA}   & cot & gen  & orig  & 10 & +.042  & --- \\
\rowcolor{tintQwen} \mQwen{Qwen-7B} & ans & logit & blank & 20 & $-.042$ & +.064 \\
\rowcolor{tintLLaVA} \mLLaVA{LLaVA}   & ans & logit & blank & 20 & +.042  & $-.094$ \\
\bottomrule
\end{tabular}
\caption{Blank-image and full-CoT sanity controls ($\alpha{=}5$, L16).
Eval: gen $=$ full $128$-token generation hit rate, logit $=$ next-token
$\Delta$argmax. Img: orig $=$ true image, blank $=$ uniform-grey canvas.
Blank-image patching collapses the original-image effect to within
noise; the full-CoT rows are intentionally underpowered ($n{=}10$/q)
and cross-checked by Table~\ref{tab:stepwise}.}
\label{tab:blank}
\end{table}

\section{Qwen-32B Scaling Sanity}
\label{sec:appendix:qwen32b}

Qwen-32B has $64$ layers, more than twice Qwen-7B's $28$, so the
late-emergence pattern observed at L16 in Qwen-7B is expected to shift to
proportionally deeper layers. This appendix performs the sanity check.

Table~\ref{tab:qwen32b} reports a layer sub-sweep on $n{=}30$/q and a
larger $n{=}168$ rerun at L32. CoT argmax stays at $0.000$ at every
layer tested, while answer-only $\Delta$argmax grows monotonically from
$+.019$ at L16 to $+.089$ at L32, with the $n{=}168$ rerun confirming
$+.085$ $[.060,.111]$. This mirrors Qwen-7B's late-emergence shape at a
deeper layer index, supporting the scaling interpretation of
\S\ref{sec:transport:late}.

The sweep also clarifies Qwen-32B's transport pattern. Within Qwen-7B
the obj\_word route is open under both prompts (the open pattern),
but Qwen-32B's CoT $\Delta$argmax is null even at the deepest answer-only
peak. The transition from 7B to 32B therefore moves the model from the
open pattern into the prompt-selective pattern identified in
\S\ref{sec:variation:patterns}. The Qwen family is the only family
where the transport pattern is not stable across scales, providing a
within-family counter-example to treating family as a sufficient predictor
explicitly flag in the main text.

\begin{table}[!tbp]
\centering
\footnotesize
\setlength{\tabcolsep}{3pt}
\begin{tabular}{llrrrr}
\toprule
Run & Pmt & Lyr & $\Delta_{\mathrm{argmax}}$ & 95\% CI & gain \\
\midrule
\multicolumn{6}{l}{\textit{$n{=}30$/q layer sub-sweep}} \\
\rowcolor{tintQwen}     & cot & 16 & $-.003$ & --- & +.057 \\
\rowcolor{tintQwen}     & cot & 24 & +.003  & --- & +.095 \\
\rowcolor{tintQwen}     & cot & 32 & +.003  & --- & +.108 \\
\rowcolor{tintQwen}     & ans & 16 & +.019  & --- & +.165 \\
\rowcolor{tintQwen}     & ans & 24 & +.036  & --- & +.360 \\
\rowcolor{tintQwen}     & ans & 32 & +.089  & --- & +.676 \\
\midrule
\multicolumn{6}{l}{\textit{$n{=}168$ rerun at L32}} \\
\rowcolor{tintQwen}     & cot & 32 & +.000 & $[.000,.000]$ & +.089 \\
\rowcolor{tintQwen}     & ans & 32 & +.085 & $[.060,.111]$ & +.721 \\
\bottomrule
\end{tabular}
\caption{Qwen-32B obj\_word $\Delta$argmax and target logit gain across
layers (top: $n{=}30$/q sub-sweep without CIs; bottom: $n{=}168$ rerun
with 95\% bootstrap CIs).
Answer-only $\Delta$argmax grows monotonically from L16 to L32 while
CoT stays at zero, mirroring Qwen-7B's late-emergence pattern at a
deeper layer index.}
\label{tab:qwen32b}
\end{table}

\section{Implementation Details}
\label{sec:appendix:impl}

This section documents the implementation choices that the main text
references but does not spell out: how the object-word position is
located in each tokenizer, how centroids are constructed without
in-fold leakage, how the bootstrap is parametrized, where the patching
hook is inserted, and what the compute budget was.

\paragraph{Object-word position rule.}
For each MCQ sample, we tokenize the question with the model's own tokenizer
and locate the contiguous substring matching the referring expression for
the gold object. The patch is applied to all tokens in this matched span.
If an exact span match is unavailable, we fall back to the matched content
word anchor tokens after removing common stopwords. This is the
implementation used for the obj\_word cells and matches the token-span
rule used by the evaluation scripts.

\paragraph{Centroid construction.}
For sample $s$ in fold $f$, we compute heldout centroids
$C^{(\ell)}_q$ from folds $\{1,\ldots,5\}\setminus\{f\}$ with gold quadrant $q$.
Patching uses $\alpha(C^{(\ell)}_{\text{target}}-C^{(\ell)}_{\text{source}})$,
guaranteeing no in-fold leakage.
The same object-word-derived centroids are reused for prefix\_last
interventions; no separate prefix\_last basis is fitted. This holds the
direction source fixed while changing the injection position.
Random-direction baselines sample isotropic Gaussian directions,
project to unit norm, and rescale to the L2 norm of the $\Delta$-centroid.

\paragraph{Bootstrap.}
Sample-cluster bootstrap with 2000 resamples;
the resampling unit is the sample id (not the (sample, target) pair),
so CIs reflect across-sample variability.

\paragraph{Hook implementation.}
Direction patching is a forward-pre-hook on the residual stream at the
chosen LM layer. The hook adds
$\alpha(C_{\text{target}}-C_{\text{source}})$ to the residual at every
matched object-word token (or at the final prefix token for prefix\_last).
Logits at the next-token position
are read for $\{$A, B, C, D$\}$.

\paragraph{Compute.}
All 7B--8B models run on a single A800 80GB GPU.
Qwen-32B and Gemma-3-27B use two A800 GPUs with model parallelism.
Each canonical cell (one layer $\times$ one prompt $\times$ one position)
takes ${\sim}$20--40 minutes per model; the source-sample count follows
the per-model protocol in Section~\ref{sec:setup}.
The full experimental grid (10 models $\times$ 7 prompts $\times$ multi-layer sweeps
$\times$ two positions) totals approximately 200 A800 GPU-hours.

\section{Attention Head Knockout Details}
\label{sec:appendix:knockout}

The head-knockout experiment in \S\ref{sec:variation:validation} reports a
single number: L27.H21 in InternVL3-8B recovers $+0.063$ $\Delta$argmax
under CoT. This appendix documents the full sweep used to localise that
head, so that the result is not the product of multiple comparisons or a
cherry-picked layer.

\paragraph{InternVL3-8B multi-layer scan.}
We zero each of the $32$ attention heads at each of four candidate
downstream layers (L17, L20, L24, L27) under CoT, $n{=}50$/q. For each
layer we report the best-head recovery. Table~\ref{tab:knockout_iv3}
shows that only L27.H21 exceeds the $+0.03$ significance threshold
($+0.062$ at $n{=}50$, $+0.063$ at $n{=}100$ confirmation). All other
$31 \times 4 = 124$ head--layer cells are at or below threshold. Under
answer-only the same head produces only $+0.007$, ruling out a generic
``always-helpful'' interpretation of H21 and supporting its identification
as a CoT-specific gating candidate in this architecture.

\paragraph{LLaVA-OV-7B: distributed gating.}
A nine-layer scan over L17--L27 in LLaVA, also under CoT $n{=}50$/q,
finds no head exceeding the same $+0.03$ threshold. The best candidate
is L22.H25 at $+0.040$ ($n{=}50$), which shrinks to $+0.013$ at
$n{=}100$ confirmation and remains $+0.003$ under answer-only.
LLaVA's gating is therefore distributed rather than concentrated in a
single head, consistent with the broader observation that gate
\emph{localisation}, not gate \emph{existence}, is architecture-dependent.

We do not run head knockout on Qwen-7B because it is open at
obj\_word and therefore has no gate to localise. A full
upstream--downstream circuit account, including which heads write the
gating signal and which downstream paths it suppresses, remains open and
is discussed in the Limitations section of the main text.

\begin{table}[!tbp]
\centering
\footnotesize
\setlength{\tabcolsep}{3pt}
\renewcommand{\arraystretch}{1.05}
\begin{tabular}{l r l r}
\toprule
KO Layer & Baseline $\Delta$ & Best Head & Recovery \\
\midrule
\rowcolor{gray!8} L17 & +0.072 & H14 & +0.032 \\
\rowcolor{gray!8} L20 & +0.078 & H21 & +0.028 \\
\rowcolor{gray!8} L24 & +0.080 & H18 & +0.006 \\
\rowcolor{gray!8} L27 & +0.080 & \textbf{H21} & \textbf{+0.062} \\
\midrule
\multicolumn{4}{l}{\textit{L27 confirmation ($n{=}100$)}} \\
\midrule
\rowcolor{gray!8} L27 (CoT) & +0.095 & \textbf{H21} & \textbf{+0.063} \\
\rowcolor{gray!8} L27 (AO)  & +0.236 & H21 & +0.007 \\
\bottomrule
\end{tabular}
\caption{InternVL3-8B head knockout (steer L16, obj\_word, RefCOCO, $\alpha{=}5$).
KO Layer is the layer at which a single attention head is zeroed out;
Baseline $\Delta$ is the CoT $\Delta$argmax without knockout;
Recovery is the change after knockout (positive value $=$ gate is opened);
AO $=$ answer-only.
Bold values mark the only head and condition above the $+0.03$ significance threshold.}
\label{tab:knockout_iv3}
\end{table}

\section{Responsible NLP Checklist}
\label{sec:appendix:checklist}

\paragraph{Reproducibility.}
All hyperparameters are reported: 5-fold crossfit,
$\alpha{=}5$ canonical / $\alpha{=}10$ for the prefix\_last ladder,
sample-cluster bootstrap with 2000 resamples.
Code and scripts will be released.

\paragraph{Data.}
All evaluation uses three public research datasets:
RefCOCO and RefCOCO+ \citep{refcoco2014, refcoco_yu2016},
the GQA scene-graph benchmark \citep{gqa2019},
and COCO captions \citep{coco2014} for the axis-transfer probe.
Our MCQ derivation reuses the existing bounding-box and scene-graph
annotations and introduces no new human labels. No personally
identifiable information is collected or released.

\paragraph{Compute.}
Total: ${\sim}$200 A800 GPU-hours for all experiments.

\paragraph{AI assistants.}
AI coding assistance was used for codebase management and LaTeX formatting;
all scientific claims, experimental design, and analysis are by the authors.

\paragraph{Ethical considerations.}
This paper studies model internals on a spatial MCQ task and does not
involve user data, deployed systems, or sensitive content.

\end{document}